\documentclass{article}

\usepackage[preprint]{neurips_2023}

\usepackage[utf8]{inputenc} 
\usepackage[T1]{fontenc}    
\usepackage{hyperref}       
\usepackage{url}            
\usepackage{booktabs}       
\usepackage{amsfonts}       
\usepackage{nicefrac}       
\usepackage{microtype}      
\usepackage{xcolor}         

\usepackage{amssymb}
\usepackage{graphicx}
\usepackage{amsmath}
\usepackage{algorithm} 
\usepackage{algpseudocode} 
\floatname{algorithm}{Pseudocode}

\hypersetup{colorlinks=true,allcolors=[rgb]{0,1,0}}

\title{Affinity-based Attention in Self-supervised Transformers Predicts Dynamics of Object Grouping in Humans} 

\author{
  Hossein Adeli$^{1,2*}$,\, Seoyoung Ahn$^{2}$,\, Nikolaus Kriegeskorte$^{1}$, Gregory J. Zelinsky$^{2,3}$\\ \\
  $^{1}$Zuckerman Mind Brain Behavior Institute, Columbia University, New York, USA \\ \\
  $^{2}$Department of Psychology, $^{3}$Department of Computer Science, 
  Stony Brook University, New York, USA \\ \\
    $^{*}$ corresponding author: \texttt{hossein.adelijelodar@gmail.com}\\
  }

\begin{document}

\maketitle

\begin{abstract}
  The spreading of attention has been proposed as a mechanism for how humans group features to segment objects. However, such a mechanism has not yet been implemented and tested in naturalistic images. Here, we leverage the feature maps from self-supervised vision Transformers and propose a model of human object-based attention spreading and segmentation. Attention spreads within an object through the feature affinity signal between different patches of the image. We also collected behavioral data on people grouping objects in natural images by judging whether two dots are on the same object or on two different objects. We found that our models of affinity spread that were built on feature maps from the self-supervised Transformers showed significant improvement over baseline and CNN based models on predicting reaction time patterns of humans, despite not being trained on the task or with any other object labels. Our work provides new benchmarks for evaluating models of visual representation learning including Transformers. 
\end{abstract}


\section{Introduction}

A fundamental problem that our visual system must solve is how to group parts of the visual input together into coherent whole objects \citep{peters2021capturing}. The role of attention in solving this problem has been experimentally studied for decades \citep{treisman1996binding, adeli2022brain}. A proposed solution is that attention can bind object features through activation spreading within an object using horizontal connectivity in retinotopic visual areas \citep{roelfsema2023solving}. However, the modeling work in this domain has focused on bottom-up, Gestalt cues, and clear object boundaries for how attention can spread within an object to bind its features (e.g. the "growth cone" model \citep{jeurissen2016serial}). A compelling model of primate vision, however, should be able to handle natural images where object boundaries are frequently ambiguous and bottom-up cues must be combined with prior object-specific knowledge. 

\begin{figure}[h]
\begin{center}
\includegraphics[width=0.35\linewidth]{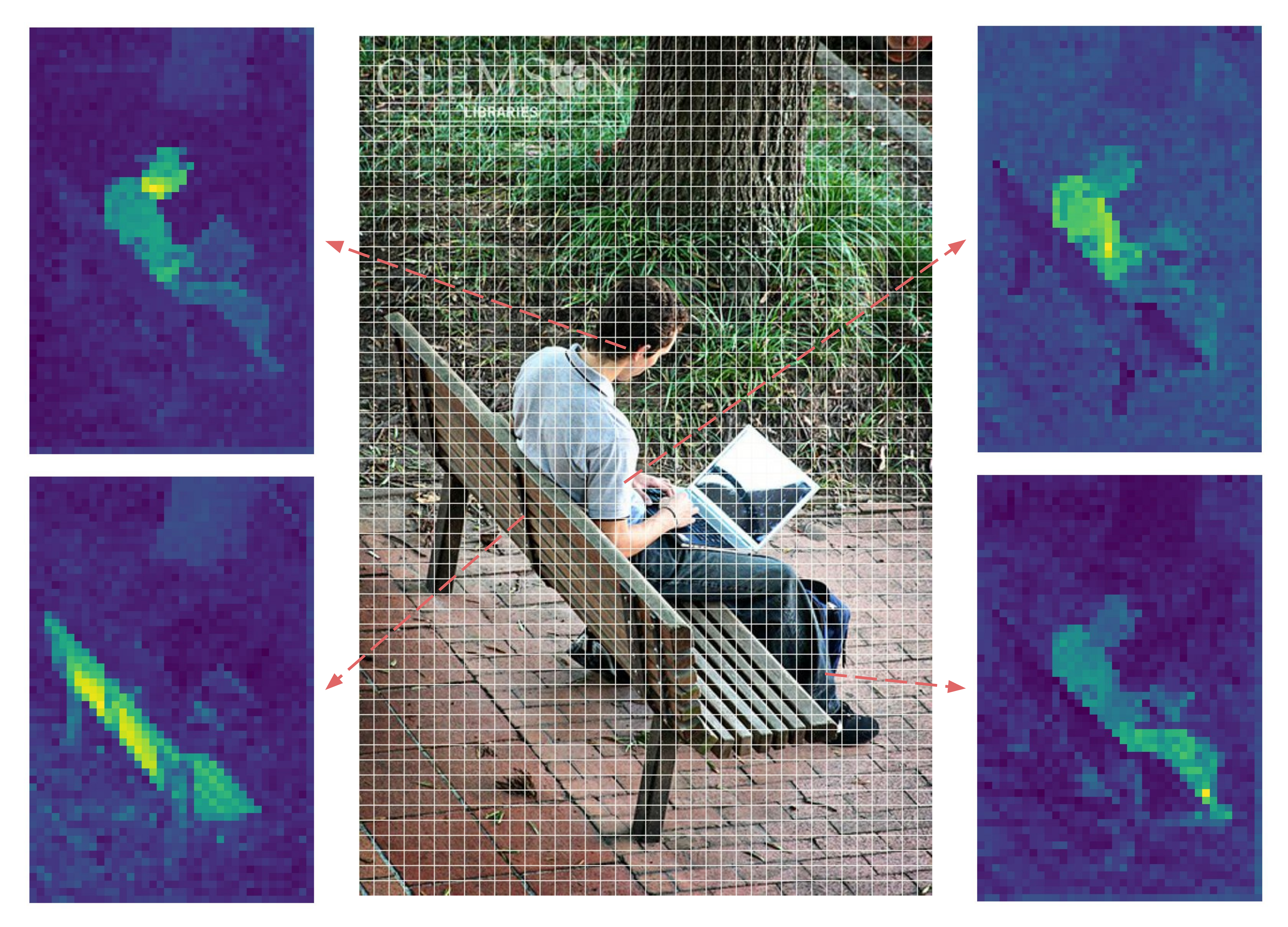}
\end{center}
\caption{Example affinity maps for a few image patches from the grid, generated using features from DINO.} 
\label{fig:affinity_maps}
\end{figure}

Building a model of object-based attention applicable to natural images requires the modeling of connectivity between image regions that can guide the spread of attention. In this work, we test the hypotheses that features from recent vision Transformers can capture this connectivity and are therefore well-suited to address the spreading of object-based attention. To that end, we introduce a model of object-based attention built on self-supervised vision Transformers. In this model, feature similarly, which we call affinity (Fig.~\ref{fig:affinity_maps}) \citep{chen2022unsupervised}, guides the spread of attention in the visual input to perform perceptual grouping, playing the role of long-range lateral connections in linking distance points of the visual input in the retinotopic maps of the ventral pathway \citep{gilbert_top-down_2013, ramalingam2013top}. We also designed and collected human data on a well-controlled behavioral experiment to probe how people group complex objects in natural scenes. We then test the models on two benchmarks based on this dataset. The first benchmark evaluates the alignment between the affinity signals in the feature maps and the ground truth object boundaries, measuring the extent to which the learned representations are object-centric. The second benchmark evaluates the models' ability to predict human object-based perception and behavior, measured by an object grouping task.



\section{Related works}

\paragraph{Models of object-based attention and grouping in the brain:} The human visual system continuously captures objects in a scene by spatially segregating their features from the background and dynamically grouping those features. Formation of these objects rely on different grouping signals ranging from part-whole matching and Gestalt processes to prior semantic knowledge of object categories \citep{greff2020binding, vecera2000toward, wagemans2012century, gilbert_top-down_2013}. Most modeling work, however, has been focused on understanding how the former, more bottom-up cues can be implemented in the retinatopic maps of the visual cortex. For example, the models of "association fields" \citep{field1993contour} suggests the effective connectivity between units depends on the similarity between the simple represented features (e.g. orientation). Attention is believed to modulate the effective connectivity between the units and therefore guide the grouping process. Most modeling work in this domain has also focused on  Gestalt cues and clear object boundaries \citep{jeurissen2016serial}. These models therefore cannot be applied to objects in natural context where objects become less spatially separated or when contours lack definition.

\paragraph{Self-supervised Vision Transformers:}
Application of Transformers to vision has been very successful, with these models outperforming convolutional neural networks (CNNs) on object recognition and other tasks \citep{dosovitskiy2020image}. In Transformers, the visual input is first divided into different patches that are then encoded as feature vectors called tokens. At each layer of processing, a given token, that represents a particular image patch, updates its value by interacting with and mixing ("attending" to) the values of all other tokens that it finds relevant. The selective nature of this mixing has motivated naming this process "attention" in Transformers \citep{vaswani2017attention}. These attention weights in supervised vision Transformers have been shown to perform some perceptual grouping \citep{mehrani2023self}. More recently, studies have explored training these models on self-supervised objectives yielding some intriguing object-centric properties that are not as prominent in the models trained for classification. When trained with self-distillation loss (DINO, \cite{caron2021emerging} and DINOv2 \cite{oquab2023dinov2}), the attention values contain explicit information about the semantic segmentation of the foreground objects and their parts, reflecting that these models can capture object-centric representations without labels. In Masked auto-encoding (MAE, \cite{he2022masked}), the input image is significantly occluded and the model is trained to reconstruct the whole image from a small number of visible patches. Minimizing the reconstruction loss enables the model to learn object-centric feature that yield great performance on other downstream tasks. 


\paragraph{Self-supervised Transformers for objects and part discovery:}
There have been recent attempts to investigate the extent to which self-supervised Transformers can learn high-level characteristics of a scene. These studies involve computing feature similarity among all tokens and examining their correspondence with high-level concepts such as objects and parts. LOST \citep{simeoni2021localizing} and Tokencut \citep{wang2022self} use the similarity graph to perform unsupervised object discovery showing success when there is one salient object in the scene. Other work \citep{amir2021deep} have used the feature similarity to perform co-segmentation of object parts. These results collectively corroborate that vision Transformers trained with a self-supervised objective begin to represent object-centric information, meaning the patches that have the highest affinity to a given patch are likely to be on the same object (Fig.~\ref{fig:affinity_maps}).


\section{Behavioral experiment}

\begin{figure}[ht]
\begin{center}
\includegraphics[width=\linewidth]{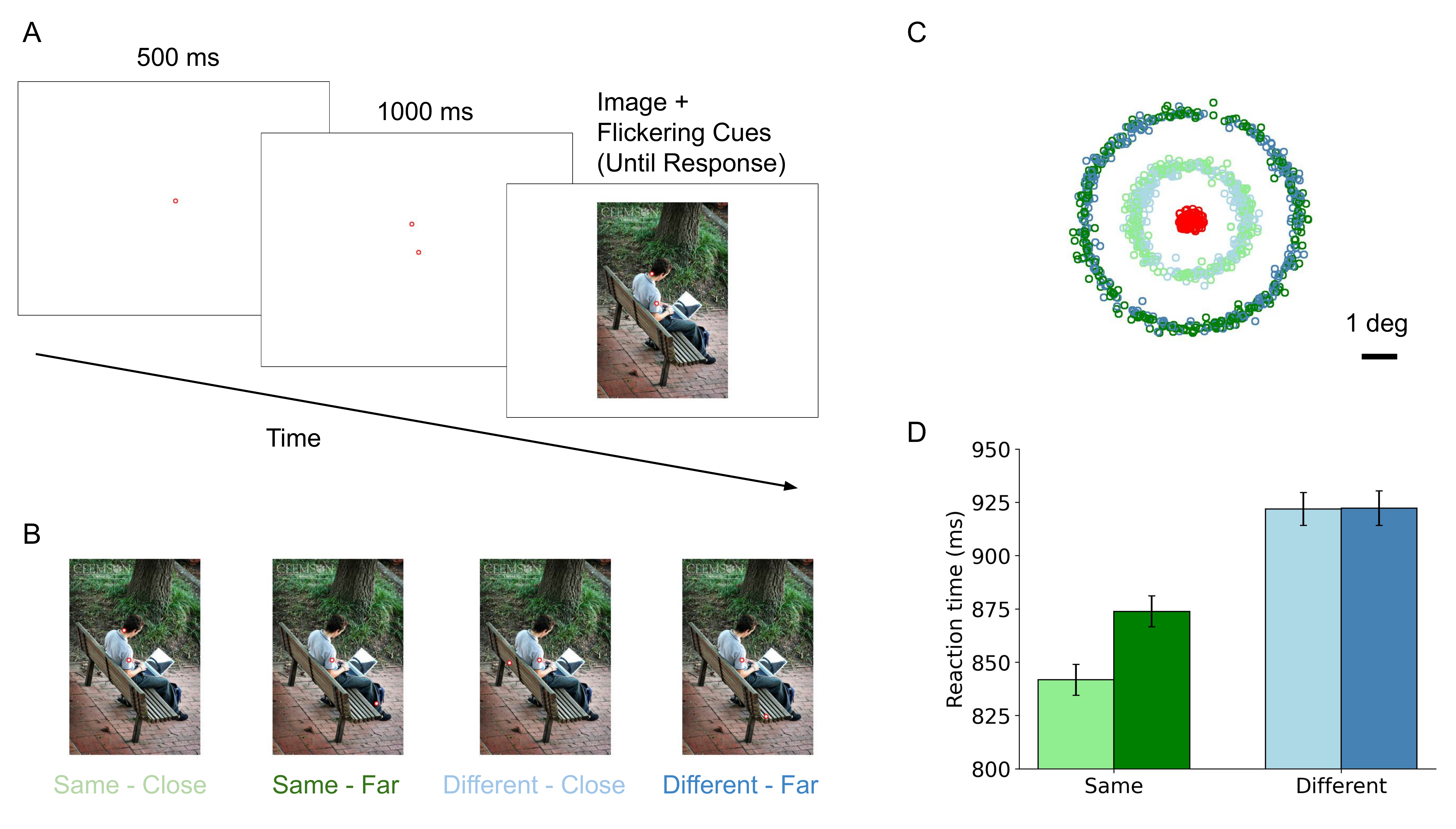}
\end{center}
\caption{\textbf{A)} Behavioral procedure. Participants maintain fixation on the center dot during the trial. \textbf{B)} Sample trial from all four conditions, each coded by color. \textbf{C)} Placement of dots in all conditions and trials. \textbf{D)} Mean reaction time for correct trials by condition, with SEM.} 
\label{fig:beh_exp}
\end{figure}

We use a "two-dot" paradigm (Fig.~\ref{fig:beh_exp}) to directly probe how humans group and segment the regions of natural images into objects. In this paradigm, two dots are placed on an image and participants are asked to indicate by button press whether they are on the same object or two different objects (Fig.~\ref{fig:beh_exp}A). One of these dots is always at the center location, and the other is at a peripheral location. The reaction time (RT) of this button press is the primary measure in this task, and reveals the difficulty of object grouping and the spread of attention within an object. Previous works using this paradigm have been limited in scale or have focused on simpler stimuli \citep{vecera2000toward, kim2019disentangling, korjoukov2012time}. For example, in \cite{korjoukov2012time} 24 images were hand selected to depict two instances of either a vehicle or an animal. Our work significantly scales up this effort, and our dataset is available at \href{https://github.com/Hosseinadeli/affinity_attention}{github.com/Hosseinadeli/affinity\_attention}.


\subsection{Behavioral methods}

\paragraph{Participants:}
72 undergraduate students participated in our experiment for course credit. Their mean age was 20.4 years (range = 17–32) and all had normal or corrected-to-normal vision. This study was approved by the school Institutional Review Board. 


\paragraph{Stimuli and Apparatus:}
We selected 288 images from the Microsoft COCO (Common Objects in Context) dataset (2017 validation set), which has images of complex everyday scenes depicting common objects in their natural context \citep{lin2014microsoft}. The images also come with object-level segmentations, which we used to generate four versions of each display (Fig.~\ref{fig:beh_exp}B): "same-close" (two dots on the same object with a close distance), "same-far" (two dots on the same object with a far distance), "different-close" (two dots on two different objects with a close distance), and "different-far" (two dots on two different objects with a far distance). We ensured that the distances are controlled between the two same/different conditions, thus preventing the participants from making the same/different decision based on distance. Fig.~\ref{fig:beh_exp}C shows the placement of the dots across all four conditions. Each participant saw each image only in one condition. The assignment of images to the four conditions was counterbalanced across participants so that every 4 participants saw the full set of trials. The experiment was conducted on a 19-inch flat-screen CRT ViewSonic SVGA monitor with a screen resolution of 1024×768 pixels and a refresh rate of 100 Hz. Participants were seated approximately 70 cm away from the monitor, which resulted in the screen subtending a visual angle of 30$^\circ$ ×22$^\circ$. This meant that around 34 image pixels spanned approximately 1 degree of visual angle, making the 'close' peripheral dot in the experiment located around 3 degrees from the fixation point (central dot) and 'far' peripheral dot located 6 degrees from the fixation point. Gaze position before the button response was recorded using an EyeLink 1000 eye-tracking system (SR Research) with a sampling rate of 1000 Hz. Gaze coordinates were parsed into fixations using the default Eyelink algorithm, which employed a velocity threshold of 30 degrees per second and an acceleration threshold of 8000 degrees per second squared. Calibration drift was checked before every trial, and recalibration was performed if necessary to ensure accurate eye-tracking data. 

\paragraph{Procedure:}
Participants were instructed to determine whether the two dots were on the same object or two different objects. Each trial started with the presentation of a fixation cross for 500 ms, indicating the location of the central dot. Both the central and peripheral dots were then displayed for 1,000 ms without the image. Next, the cues were superimposed on the image and flickered at a frequency of 5 Hz to ensure their visibility. During the trial, participants were required to maintain their gaze on the center dot for the entire duration. If their gaze deviated more than 1 degree of visual angle away from that location, the trial was terminated. 7 percent of the trials were removed due to breaking fixation. To record their responses, participants utilized a Microsoft gamepad controller, with buttons assigned to the "same" or "different" condition. The assignment of the same button were randomized to the right or left hand across participants to ensure the RT differences were not due to the dominant hand bias. Each participant performed 32 practice trials and 256 experimental trials. The experimental trials were divided into four blocks, with breaks provided between the blocks. The order of image presentation in each block was randomized across the trials. To provide accuracy feedback, a sound was used to indicate an incorrect response to participants. We removed one experimental image from our analyses based on the ground truth response being ambiguous, leavening 255 experimental images and 1020 (255$\times$4) trials for behavioral analyses and model evaluations.  


\subsection{Behavioral results}

\begin{figure}[ht]
\begin{center}
\includegraphics[width=\linewidth]{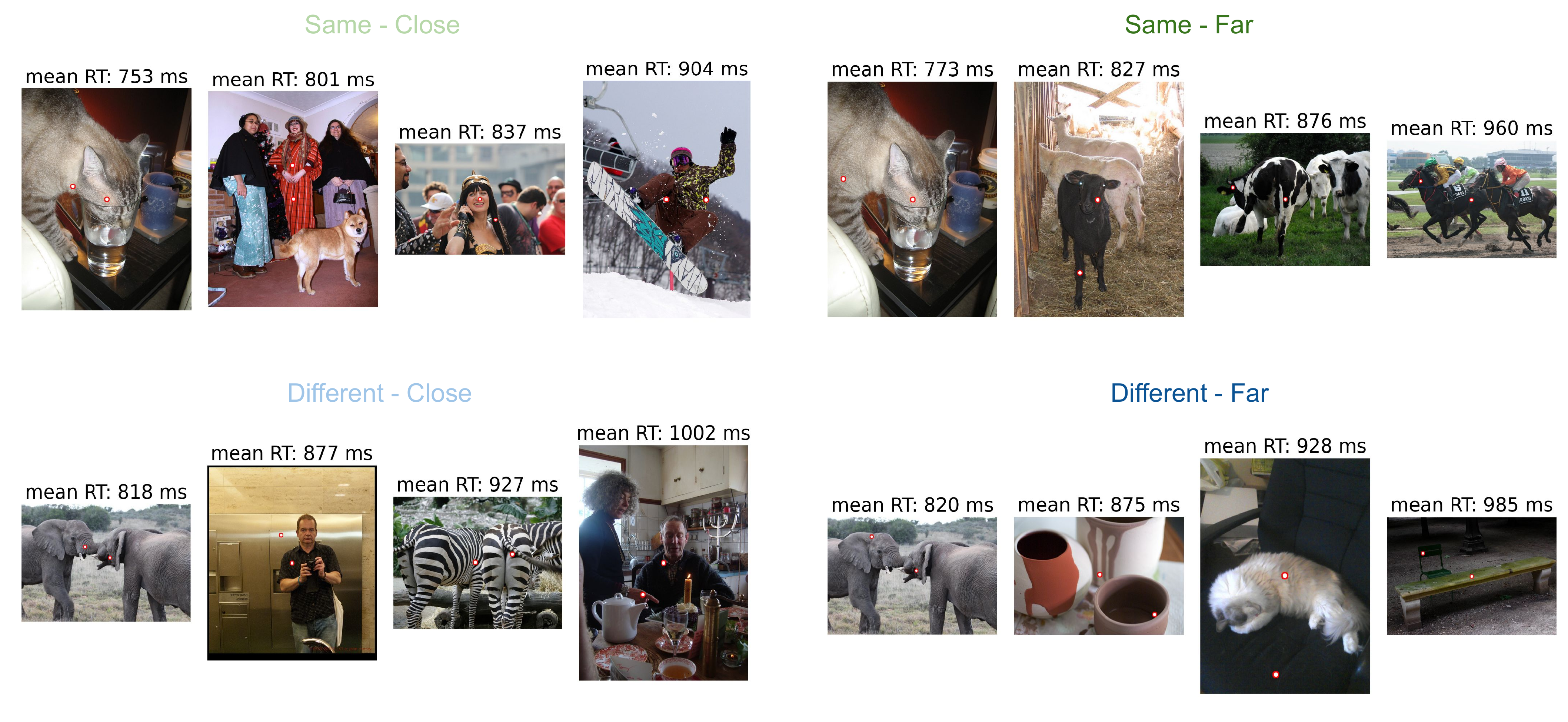}
\end{center}
\caption{\textbf{A)} Sample trials for the four conditions in our experiment. The corresponding average reaction times from subjects is displayed on top.} 
\label{fig:sample_trials}
\end{figure}

The average subject accuracy on this task was 90 percent. We only show the analyses for trials where the subject response was correct (however the patterns were largely the same when considering both correct and incorrect trials). Fig.~\ref{fig:beh_exp}D shows the RT data for each condition. Subjects were faster to respond when the two dots were on the same object compared to when the peripheral dot was on a different object. This effect is know as the same object advantage \citep{egly1994shifting}, indicating that the first dot facilitates the selection of the whole object. This effect interacted with dot distance, where we observed the fastest RTs in the close-separation same-object condition. This behavioral pattern is consistent with the hypothesis that attention spreads from the center dot within the cued object, thereby reaching the closer dot faster than the farther dot. If the peripheral dot is on a different object, dot separation would not be expected to play a large role on RTs \citep{roelfsema2023solving}. 

Fig.~\ref{fig:sample_trials} shows four sample trials for each condition with comparable difficulties. For this visualization, We first ordered the trials for each condition by their RTs and then selected the 50th, 100th, 150th and 200th trials, displayed from left to right, respectively, for each condition. Of note, RTs increase with the distance between the dots when the dots are on the same object. Comparing the cat figures (top row) illustrates this pattern for dots on the same object. When the dots are on different objects, there is no effect of dot separation on RTs, as can be seen by comparing the elephants. While average and cross condition behavioral patterns are interesting, there is also interesting variability within each condition. Task difficulty, and RTs, increase when there are within object boundaries between the dots, when the dots are on different object parts (with different textures), when the dots are on the narrower parts of the object, when they are close to the boundaries, or when there are multiple objects from the same category. Our behavioral dataset is therefore rich in capturing the variable conditions under which humans group objects in natural scenes. We will test models on how well they can predict the mean RT of the subjects for each trial. 

\section{Modeling experiments}

\begin{table}[h]
\caption{All model run specifications}
\label{tab:all_runs}
    \centering
\def\arraystretch{1.6}
\begin{tabular}{|p{3cm}||p{1.4cm}|p{1.7cm}|p{1.cm}|p{1cm}|p{3cm}| }
 \hline
  run name & training objective  & architecture & model size & patch size & feature type \\
 \hline
 DINOv2\_ViTb14 & DINO v2& ViT & base & 14 & key, query, or value\\
 \hline
 DINOv2\_ViTl14 & DINO v2& ViT & large & 14 & key, query, or value\\
 \hline
 DINOv2\_ViTg14 & DINO v2& ViT & giant & 14 & key, query, or value\\
 \hline
 DINO\_ViTb16 & DINO & ViT & base & 16 & key, query, or value\\
 \hline
 DINO\_ViTb8 & DINO & ViT & base & 8 & key, query, or value\\
 \hline
 MAE\_ViTb16 & MAE & ViT & base & 16 & key, query, or value\\
 \hline
 DINO\_ResNet50 & DINO & ResNet50 & & & conv\\
 \hline
 ImageNet\_ResNet50 & ImageNet & ResNet50 &  &  & conv\\
 \hline
\end{tabular}
    \centering
\end{table}

In our modeling experiments, we set out to evaluate a wide range of representation learning models, as shown in Table ~\ref{tab:all_runs}. We considered both Transformer and CNN based models. For the Transformer models we focused on the ViT architecture \citep{dosovitskiy2020image} with different number of parameters (base, large, or giant) using different patch sizes (8, 14, or 16). The models are trained with either the self-distillation method, DINO \citep{caron2021emerging} or the updated method, DINO version 2 \citep{oquab2023dinov2}. Also included are models trained using masked autoencoding (MAE,\citep{he2022masked}). In Transformers, each token is represented with three vectors: key, query, and value. The affinity between tokens can be calculated using any of these feature representations (self-attention is the dot product of one token's key with another token's query). For these models, we extract the patch features from the last Transformer layer and calculate affinity by performing the dot product of each token's feature with all the other tokens. Sample affinity maps for a few patches are shown in Fig.~\ref{fig:affinity_maps}, generated using the key features from the a model trained with DINO objective. Our code is available at \href{https://github.com/Hosseinadeli/affinity_attention}{github.com/Hosseinadeli/affinity\_attention}.

We also included two CNN models in our comparison, both using the ResNet50 architecture \citep{he2016deep}. One is pre-trained on ImageNet for object classification and the other is trained using the DINO method. Convolutional features are extracted from the third convolutional block of the model to have feature dimensions comparable to a Transformer model with the path size of 16. The feature tensors (with size h x w x d) are then divided into h x w feature vectors of length d to represent different patches of the image. We then take the dot product of each feature vector with all the others to calculate the affinity map for each patch. 



\subsection{Quantifying the object-centric component of affinity}

\begin{figure*}[h]
\begin{center}
\includegraphics[width=\linewidth]{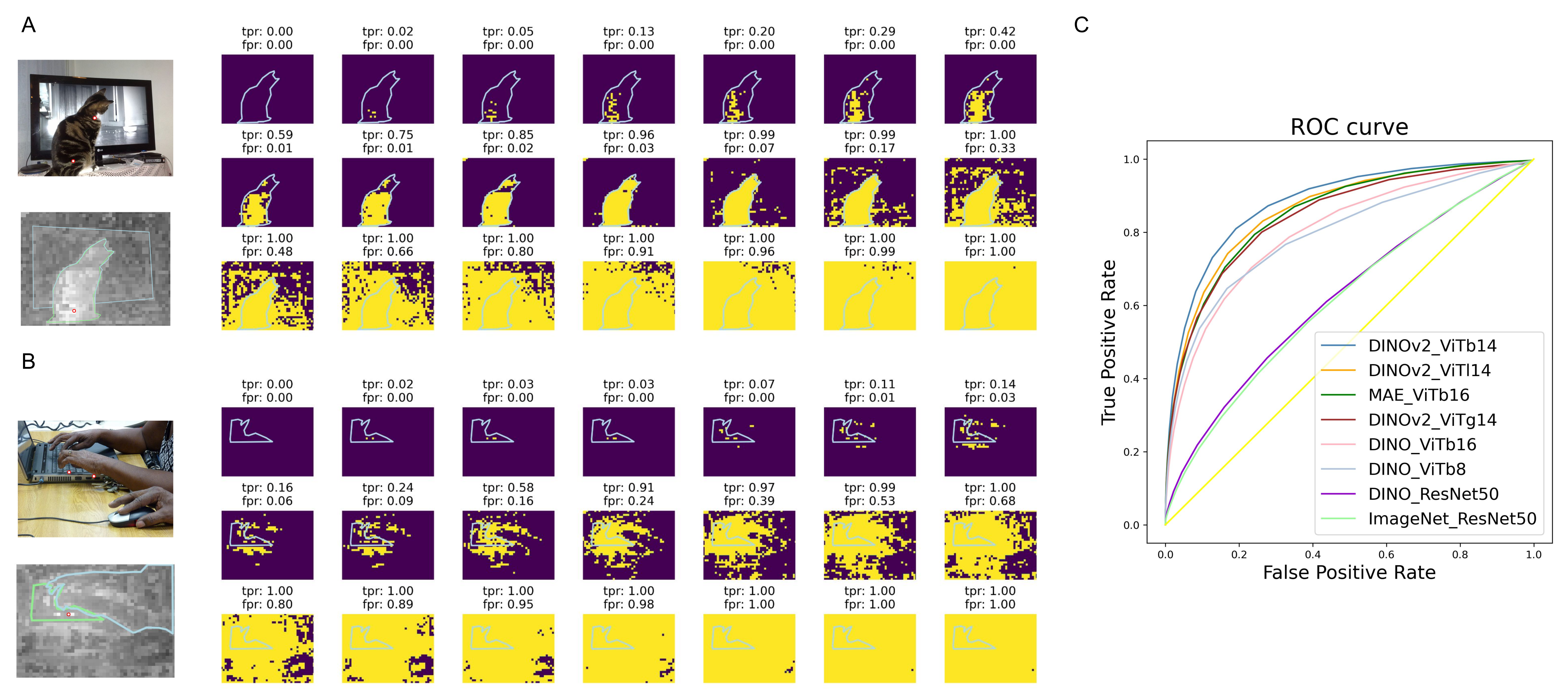}
\end{center}
\caption{\textbf{A)} A sample experimental trial with the central and peripheral dots shown on the top left. The affinity map for the peripheral dot is shown below the trial image. The activity remaining on the affinity map after decreasing a threshold in steps of 0.5 from 1 to 0. The True Positive Rate (TPR) and False Positive Rate (FPR) are displayed above the activity map for each step. The TPR increases significantly before the FPR increases, showing a strong object-centric signal in this map. \textbf{B)} Affinity map from another trial, with the same step-wise analysis of TPR and FPR as the threshold decreases. Compared to \textbf{A)}, the object-centric signal in this example is not as strong. \textbf{C)} The ROC curves for all the model runs. The figure legend is ordered by the area under the curve with model performance decreasing from top to bottom.}
\label{fig:object_centric}
\end{figure*}

Affinity between two patches of an image is driven by feature similarity. However, in order for the object-based attention to spread within an object, the affinity has to have a strong object-centric component. We quantify this by performing an ROC analyses of our experimental dataset. For each of the 1020 trials, we first extract the affinity map from the peripheral dot location. A sample map is shown in Fig.~\ref{fig:object_centric}A on the left, which used the key features from the ViT based model (trained on DINO v2 with patch size 14; DINOv2\_ViTb14 in Table ~\ref{tab:all_runs}). This affinity map represents the strength of the feature similarity between the patch at the peripheral dot location and all the other patches in the image. We utilized different thresholds to evaluate the spatial distribution of affinity signals at a specific strength and assess their alignment with the ground truth object boundary. This analysis allowed us to determine the extent to which the signals were concentrated within the object or dispersed outside of it, providing insights into the object-centric nature of the learned features. In the example shown in Fig.~\ref{fig:object_centric}A, the threshold is decreasing in steps of 0.05 from 1 to 0. True Positive Rate (TPR) is calculated for each threshold as the size of the area that is active in the object divided over the entire area of that object. False Positive Rate (FPR) is calculated as the size of the area active outside of the object divided over the entire area outside of that object. The figure shows that the TPR increases as the threshold decreases, but that the FPR does not increase, meaning the patches having the strongest affinity to the given peripheral location are in the same object. This indicates a strong object-centric signal in this affinity map. Only after the TPR reaches a high level of >0.96, does the FPR start to increase with a decreasing threshold. Figure \ref{fig:object_centric}B illustrates an example trial from the same model where the object-centric signal is not strong in the affinity map. In this case, FPR increases significantly before the TPR reaches a high level.  


Averaging the TPR and FPR across all the trials, we can then get a gist of how object-centric the affinity is for a given model. Fig.~\ref{fig:object_centric}C shows ROC curves for all the model runs in Table ~\ref{tab:all_runs}. We calculate the area under the ROC curves (AUC) as the performance measure. The figure legend is ordered from top to bottom based on decreasing AUC. For the Transformer models we report the best results among the key, query, and value features. We generally observed similar performance from the affinity maps built using the key and query features, and that they outperform the value features. This observation generally agrees with prior work on using these features for unsupervised object discovery \citep{simeoni2021localizing, wang2022self}. Results reveal that DINO v2 clearly improves upon the earlier DINO model. The MAE based affinity maps perform strongly, outperforming original DINO model but not the v2. Interestingly, the CNN-based models perform poorly on this task regardless of the training objective. The results corroborate earlier findings suggesting that self-supervised learning methods, paired with a Transformer architecture, learn more object-based features.

\subsection{Affinity spread}





\begin{figure*}[h]
\begin{center}
\includegraphics[width=\linewidth]{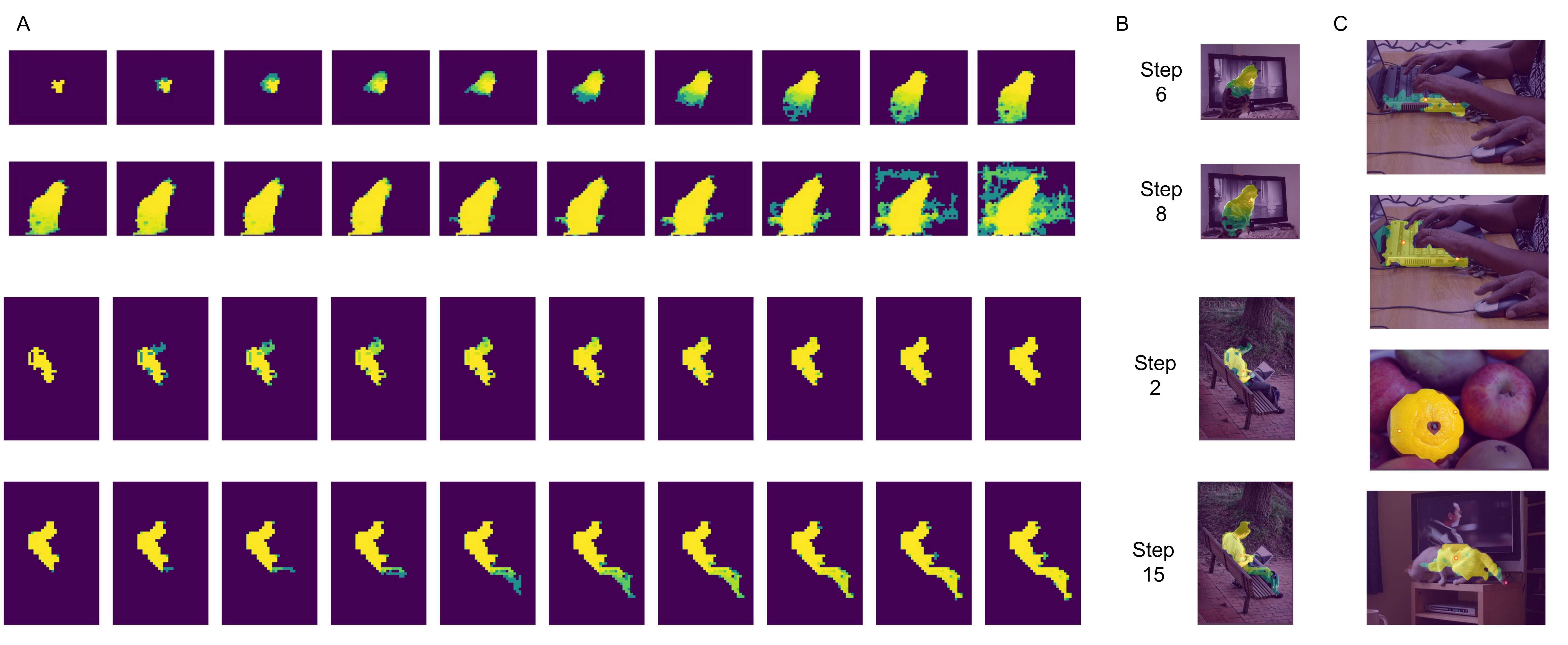}
\end{center}
\caption{\textbf{A)} 20 steps of attention spread for two trials (top vs bottom two rows), each starting from the center dot. \textbf{B)} Attention spread overlaid on the image for the steps that attention reached the peripheral dot in the close (top) and far (bottom) conditions for both trials. \textbf{C)} Examples of attention spreading in objects.}
\label{fig:spread}
\end{figure*}

We designed a simple algorithm to spread attention through an object using the affinity signal (that is normalized to have all values between 0 and 1). The model, like the subjects, starts every trial from the patch at the center dot location. From this starting location, the model then selects all the tokens with strong affinity above the starting threshold (\texttt{tau}), causing attention to spread to a bigger segment around the center dot (Fig.~\ref{fig:spread}A top-left). At each new step, the model identifies the patches in the image that have a strong affinity with the already attended segment by taking an average of the affinity maps for all the tokens over this growing segment. We put a constraint on this process to have the segment grow as one contiguous region but only allowing the newly added patches to be connected to the current segment. Fig.~\ref{fig:spread}A shows the iterative spread of attention in two sample objects over 20 steps (generated from DINOv2\_ViTb14\_k features). As the segment grows, the attention spread becomes more conservative due to the constraint placed on the segment growth and that all the patches in the segment vote where to spread next. To counter this conservative spread, the model gradually reduces the threshold (\texttt{tau\_step}) at each time step, thereby allowing attention to spread within the entire object before spreading outside of it. \texttt{Tau} and \texttt{tau\_step} are two hyperparameters in the algorithm that we explored for each run.   

The number of steps that it takes for attention to reach the peripheral dot is the model's prediction of the RT for that trial. Fig.~\ref{fig:spread}B shows the attention spread overlaid on the original images for the two trials. In both cases, the model reaches the close peripheral dot (top) in fewer steps than the far peripheral dot (bottom). Fig.~\ref{fig:spread}C shows more examples of trials where attention reaches the dot in the object. We find that, while the individual affinity maps could be noisy and scattered, the act of spreading object-based attention from a point inside the object that relies on averages of many affinity maps can yield object-centric incremental segments.



\begin{figure*}[t!]
\begin{center}
\includegraphics[width=\linewidth]{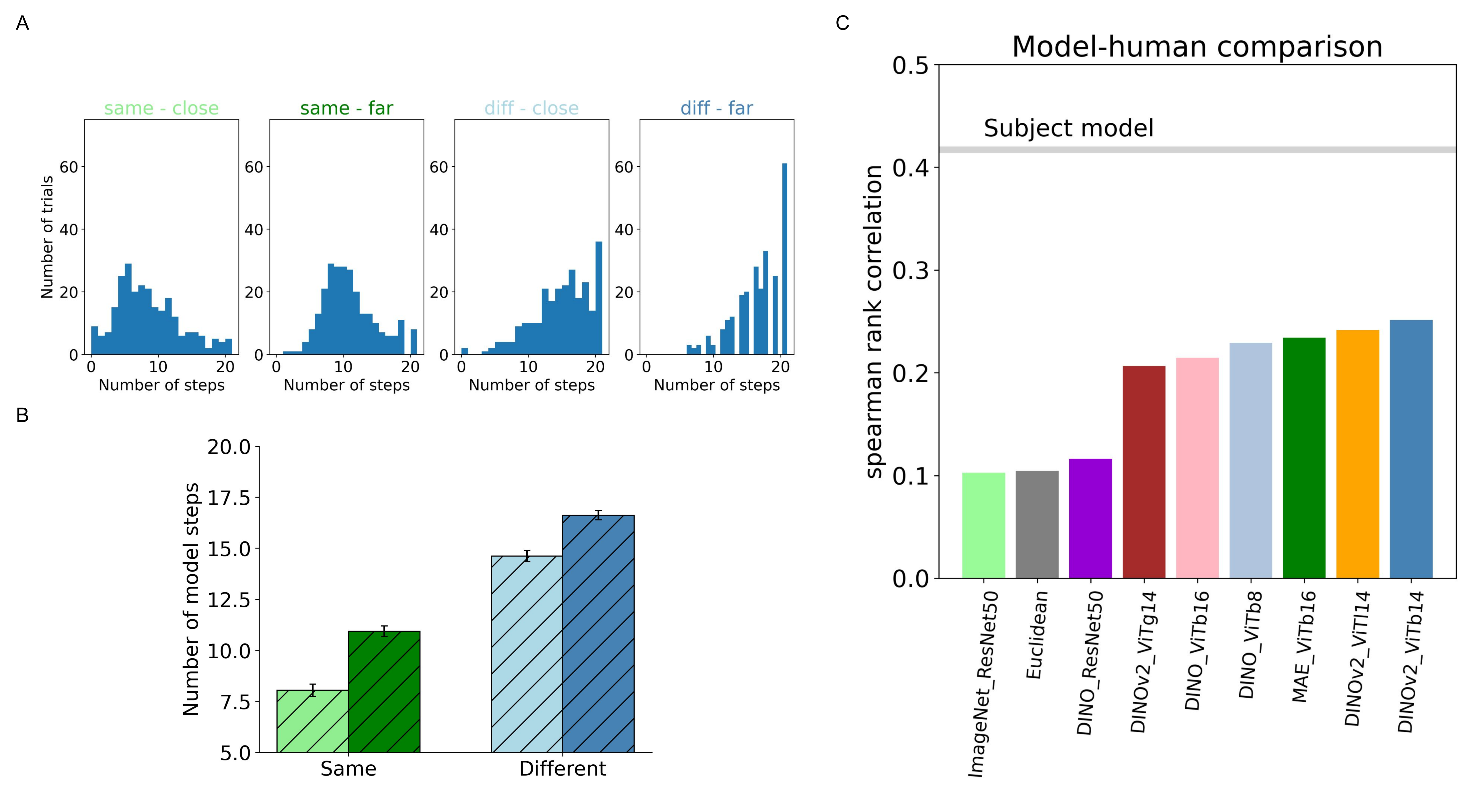}
\end{center}
\caption{\textbf{A)} Histogram of the number of trials where the attention spread reaches the peripheral dot for each condition, with number of steps on the x-axis. \textbf{B)} Mean number of model steps for attention to reach the peripheral dot in the close (light bars) and far (dark bars) conditions for both same and different trials, with SEMs. \textbf{C)} Correlation between different model predictions and average subject RT. The gray line shows the subject-subject agreement, serving as a rough upper bound on model performance.}
\label{fig:model_predictions}
\end{figure*}

Model predictions were similarly made for the "different" condition trials where two dots are located on two different objects. Given that the threshold is decreasing at each step, in some trials the attention spreads outside the object and reaches the peripheral dot even when the dot is on a different object. That step number will be the model prediction for the RT of the participants on that trial. In other cases, attention never reaches the peripheral dot when it is on a different object, in which case the model prediction would be 21 (maximum number of steps + 1). We plot in Fig.~\ref{fig:model_predictions}A the histogram of the number of trials where attention reaches the peripheral dot as a function of number of steps (1 to 21) for the DINOv2\_ViTb14 run with \texttt{tau} and \texttt{tau\_step} set at 0.8 and 0.2, respectively. The rightmost bar at bin 21 in each condition indicates the number of trials where attention did not reach the peripheral dot. Fig.~\ref{fig:model_predictions}B shows the average number of steps that the model took for attention to reach the close and far peripheral dot for each condition. The model predicts the same object advantage in the same condition, evidenced by a smaller number of steps, compared to the different condition. Also for the same condition, the model behavior shows the effect of distance on the number of steps, similar to the effect of distance on RT that we observed in humans. However, the model also showed an effect of distance in the different condition trials, contrary to the pattern in human RT. This Highlights a limitation in our approach to predicting RT in the different condition that needs to be addressed in future work. Lastly, we observed that the behavioral trends of the model shown in Fig.~\ref{fig:model_predictions}B largely stayed the same for a range of \texttt{tau} and \texttt{tau\_step} values, but that the average number of steps that it took to reach the peripheral dot was higher with higher starting thresholds and smaller steps.   

Fig.~\ref{fig:model_predictions}C shows how affinity-spread based on features from different models can predict the human behavior on the grouping task. The bars show the correlation of each model prediction with the average reaction time of the subjects across the 1020 trials. We also include a Euclidean model as a simple baseline that predicts longer RT when there is a greater distance between the two dots. The gray horizontal line labeled "Subject model" indicates the upper bound for model prediction. In order to create this model, we split the 72 subjects' data into two groups with 36 subject and correlated the average response of the two groups. We then repeated the process 50 times with different splits and then took the average of the correlation coefficients across the splits.


The results suggest that affinity spread models built using feature maps from self-supervised Transformers better align with human grouping behavior when compared to baseline models and models based on convolutional neural networks (CNN), despite not being trained on the same/different task or with any other object labels. We can observe that models exhibiting stronger object-centric affinity signals tend to achieve superior performance in predicting human behavior, a trend we expect to continue. Another notable observation is that larger models do not necessarily perform better, with ViTg14 underperforming the base and large ViT models. Given the still large gap between model and human behavior on individual trials, this model comparison and dataset will be a useful benchmark for future developments.



\section{Discussion}

In this work, we proposed a novel affinity-based model of object grouping and demonstrate that self-supervised Transformers provide a plausible mechanism for human grouping and object-based attention, thus extending the value of these models beyond core object recognition \citep{vanrullen2001time}. Our affinity spread method, building on self-supervised representation learning, does not require a large number of labeled samples for training, making this a more biologically plausible mechanism for how the primate visual system learns to group features and perceive objects. Our work also provides a new behavioral dataset and framework for evaluating models of representation learning, including Transformers. This work further contributes to computer vision by showing how object-based attention, a core element of human cognition, can be integrated into an AI model.

\bibliography{neurips_2023}

\section{Supplementary results}

In this section we provide the complete model comparison and more sample trials from our affinity-spread model.

\begin{figure}[ht]
\begin{center}
\includegraphics[width=\linewidth]{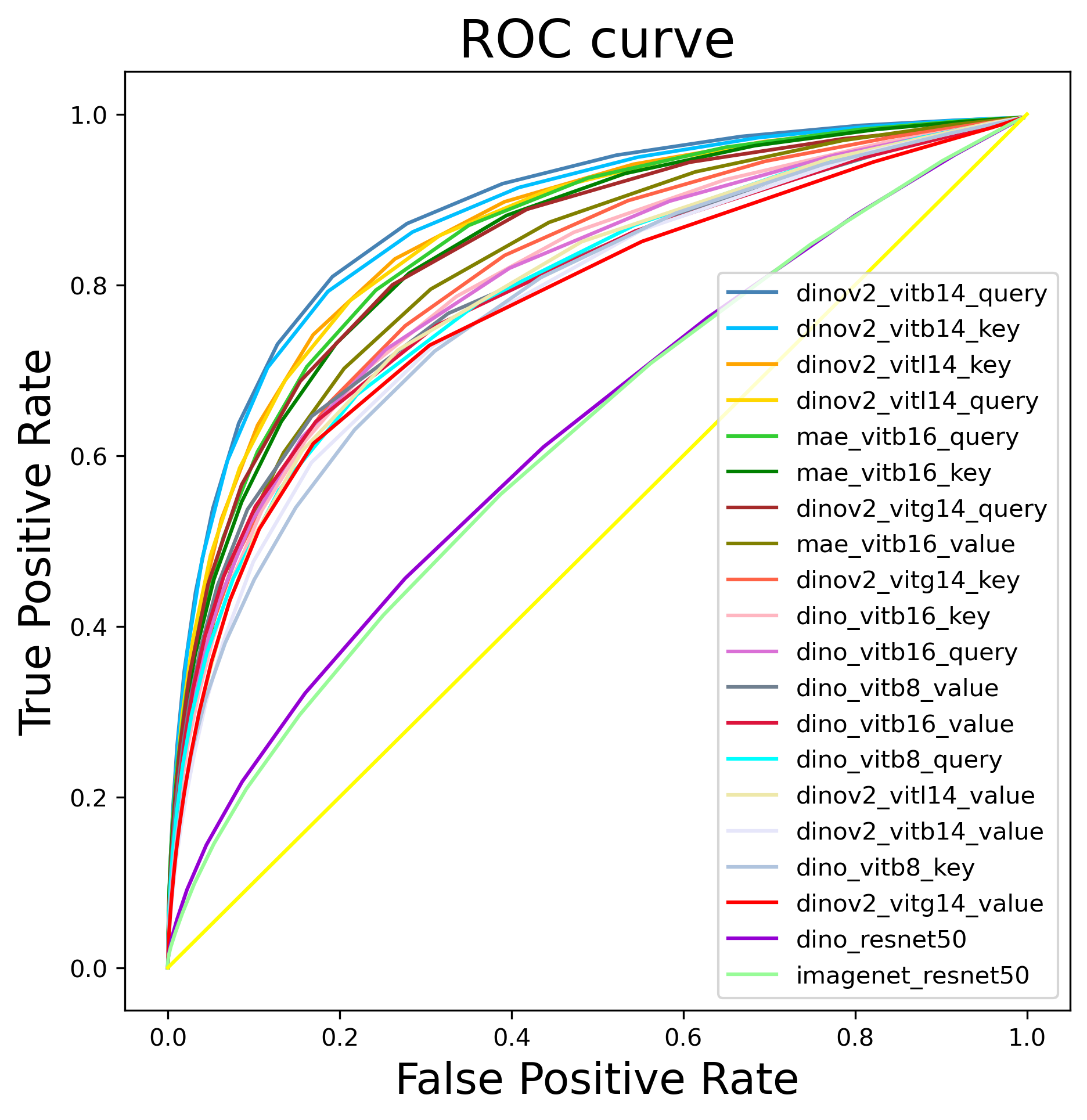}
\end{center}
\caption{ The ROC curves for all the model runs. The figure legend is ordered by the area under the curve with model performance decreasing from top to bottom.} 
\label{fig:object_centric_supp}
\end{figure}

Table ~\ref{tab:all_runs_roc} provides the Area Under the Curve (AUC) for the ROC curves of all model runs. In this table the performance is also broken down for each feature type. Fig.~\ref{fig:object_centric_supp} shows the ROC curves.

\begin{figure}[ht]
\begin{center}
\includegraphics[width=\linewidth]{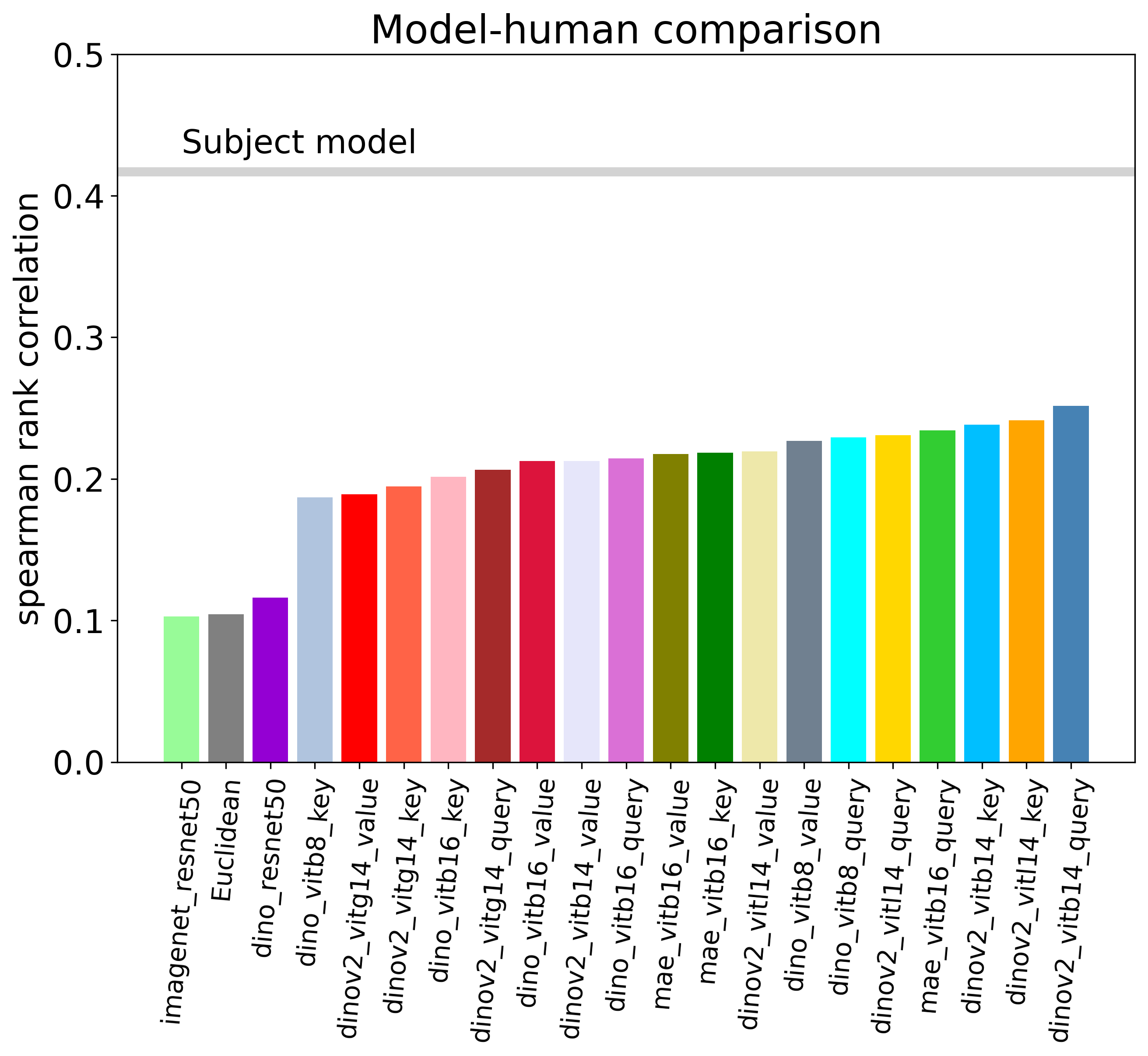}
\end{center}
\caption{Correlation between different model predictions and average subject RT. The gray line shows the subject-subject agreement, serving as a rough upper bound on model performance.} 
\label{fig:model_predictions_supp}
\end{figure}

Table ~\ref{tab:all_runs_corr} provides the correlation coefficients between all model runs and the human behavior. In this table, the performance is also broken down for each feature type. Fig.~\ref{fig:model_predictions_supp} shows this information in a bar graph.

\begin{figure}[ht]
\begin{center}
\includegraphics[width=\linewidth]{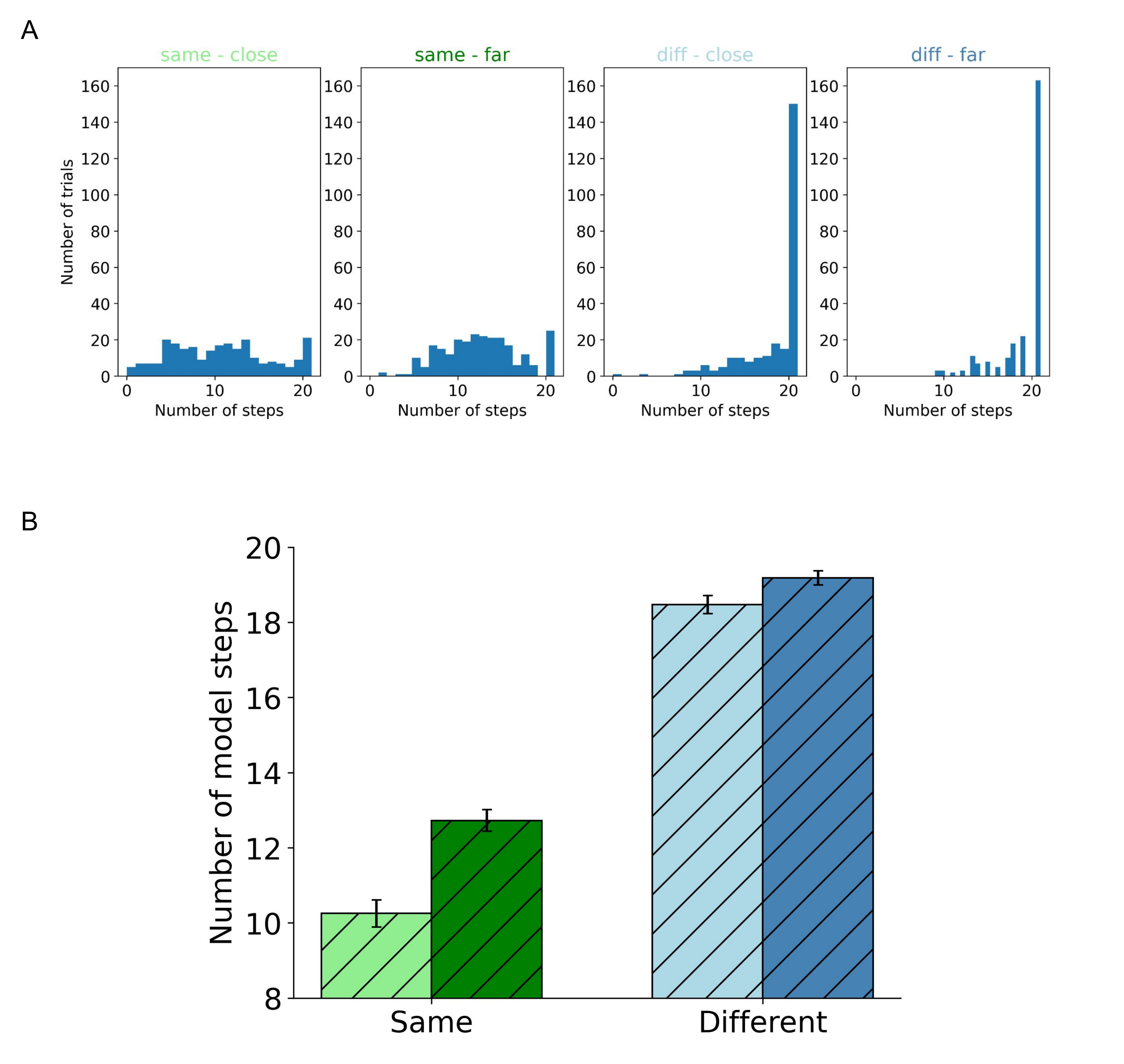}
\end{center}
\caption{ Restuls for DINOv2\_ViTb14\_query run. \textbf{A)} Histogram of the number of trials where the attention spread reaches the peripheral dot for each condition, with number of steps on the x-axis. \textbf{B)} Mean number of model steps for attention to reach the peripheral dot in the close (light bars) and far (dark bars) conditions for both same and different trials, with SEMs.} 
\label{fig:model_pred_q}
\end{figure}

Fig.~\ref{fig:model_pred_q} shows the histograms of the predictions for the best performing model. 

\begin{figure*}[h]
\begin{center}
\includegraphics[width=\linewidth]{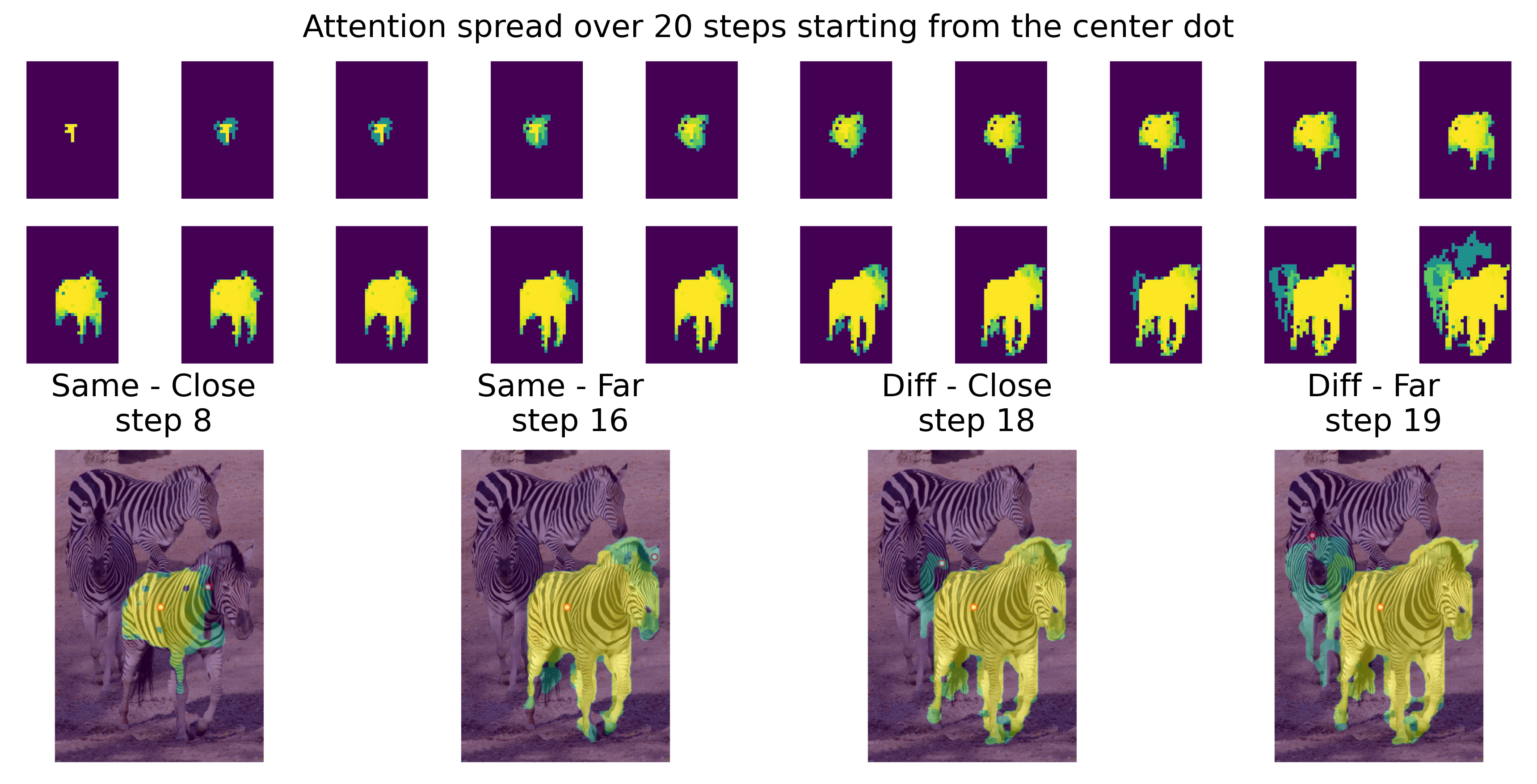}
\end{center}
\caption{20 steps of attention spreading from the center dot. Attention spread overlaid on the image for the steps that attention reached the peripheral dot in the same - close, same - far, different - close and different - far conditions shown at the bottom.}
\label{fig:spread_s1}
\end{figure*}

\begin{figure*}[h]
\begin{center}
\includegraphics[width=\linewidth]{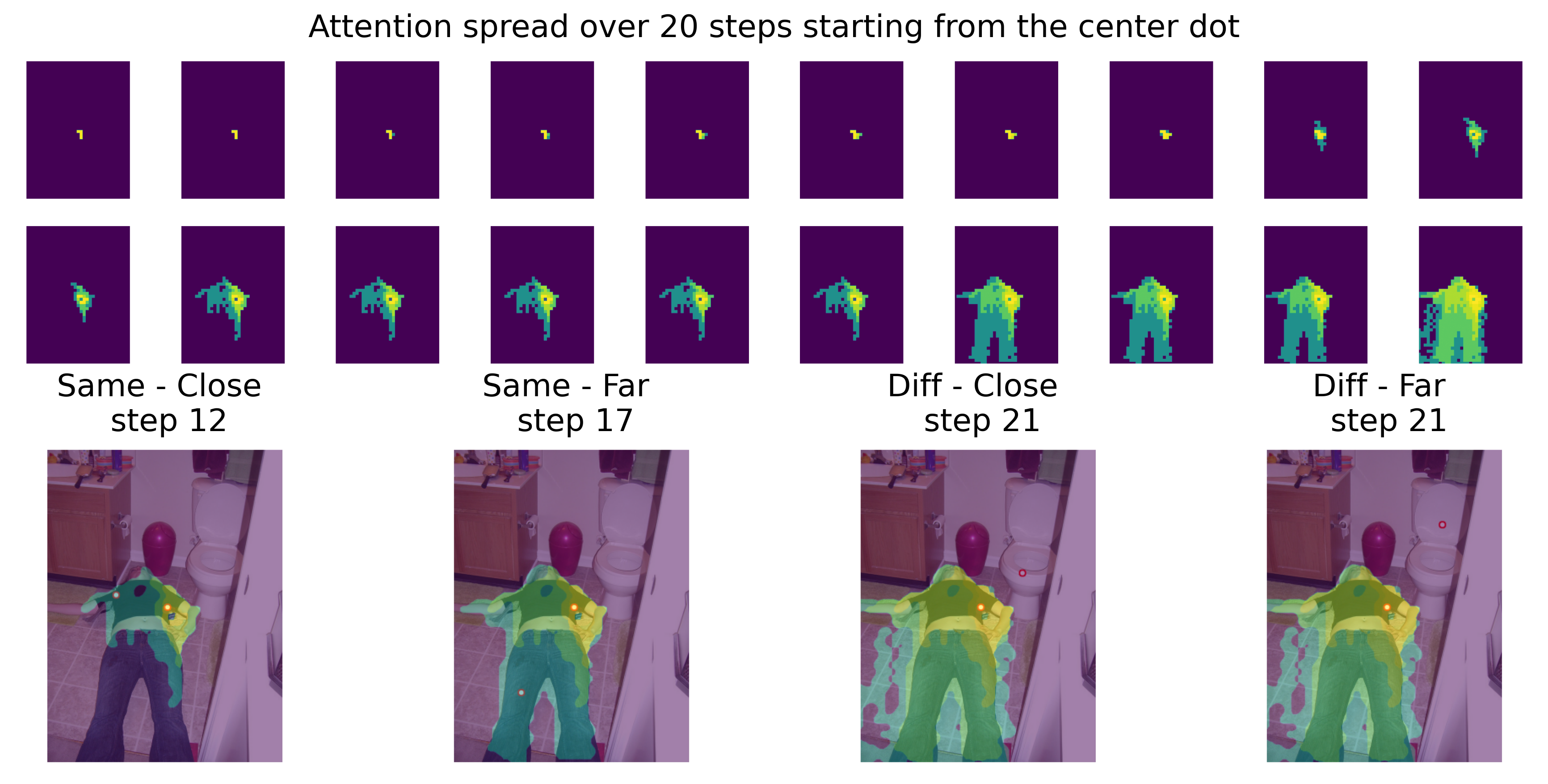}
\end{center}
\caption{20 steps of attention spreading from the center dot. Attention spread overlaid on the image for the steps that attention reached the peripheral dot in the same - close, same - far, different - close and different - far conditions shown at the bottom.}
\label{fig:spread_s2}
\end{figure*}

\begin{figure*}[h]
\begin{center}
\includegraphics[width=\linewidth]{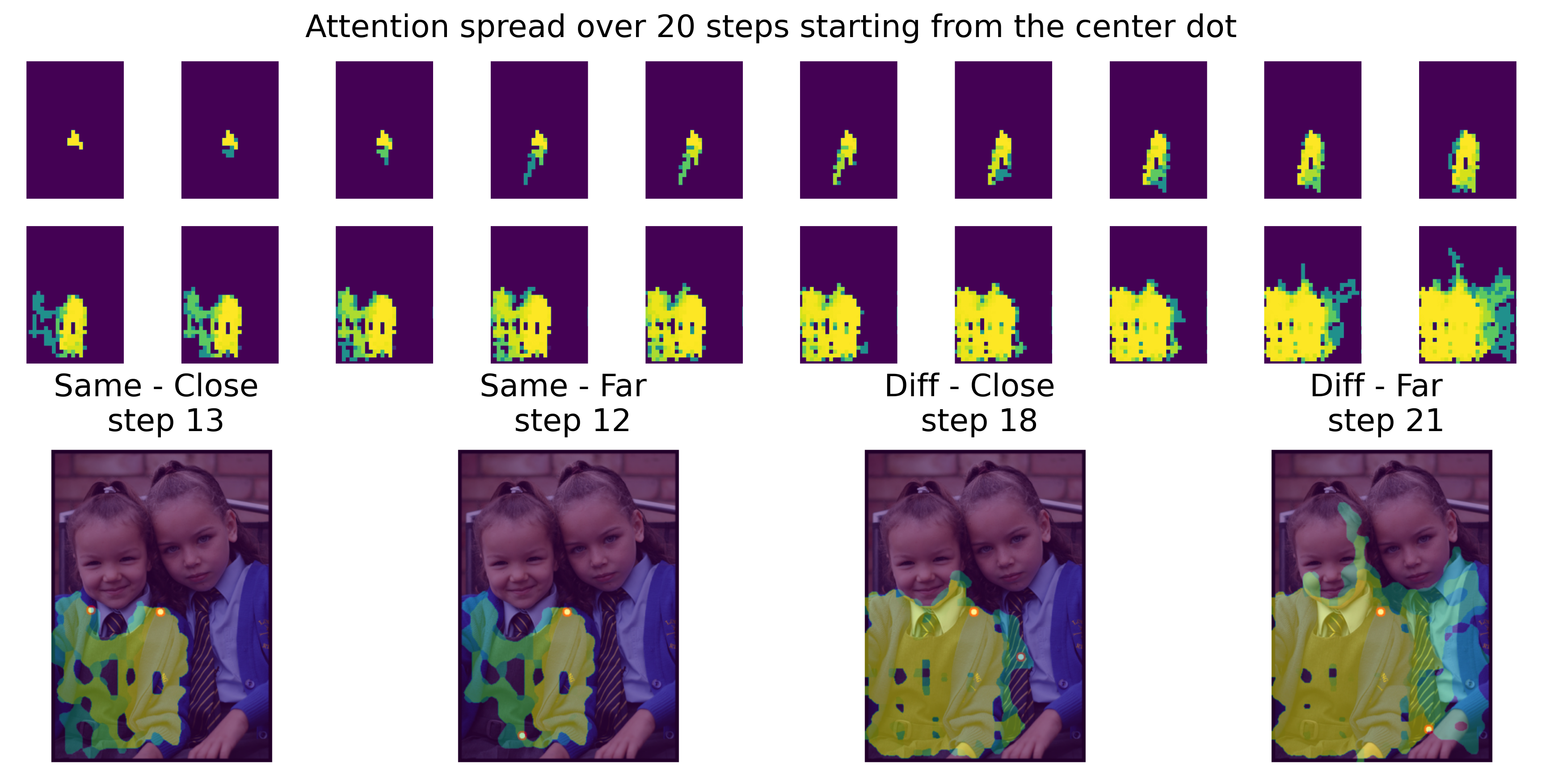}
\end{center}
\caption{20 steps of attention spreading from the center dot. Attention spread overlaid on the image for the steps that attention reached the peripheral dot in the same - close, same - far, different - close and different - far conditions shown at the bottom.}
\label{fig:spread_s3}
\end{figure*}

\begin{figure*}[h]
\begin{center}
\includegraphics[width=\linewidth]{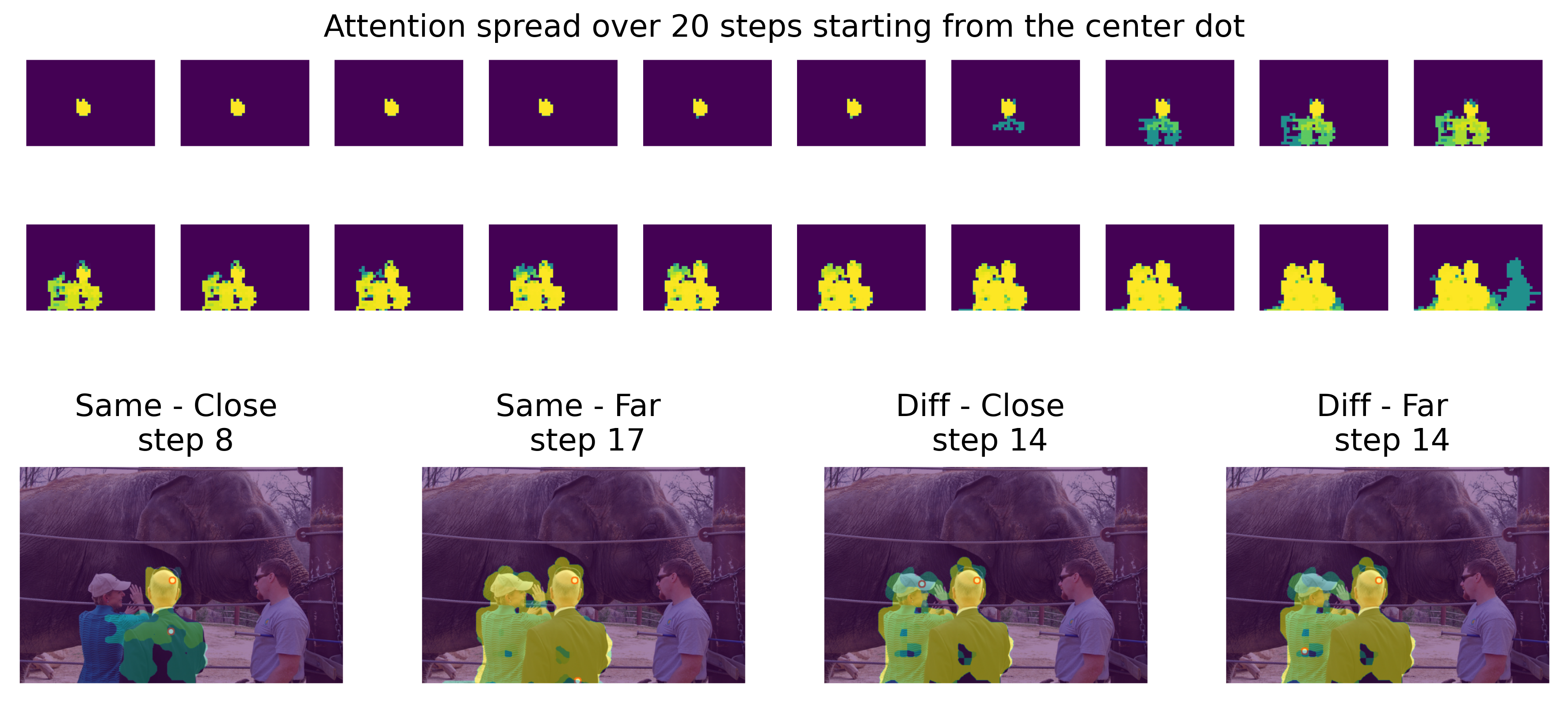}
\end{center}
\caption{20 steps of attention spreading from the center dot. Attention spread overlaid on the image for the steps that attention reached the peripheral dot in the same - close, same - far, different - close and different - far conditions shown at the bottom.}
\label{fig:spread_s4}
\end{figure*}

\begin{figure*}[h]
\begin{center}
\includegraphics[width=\linewidth]{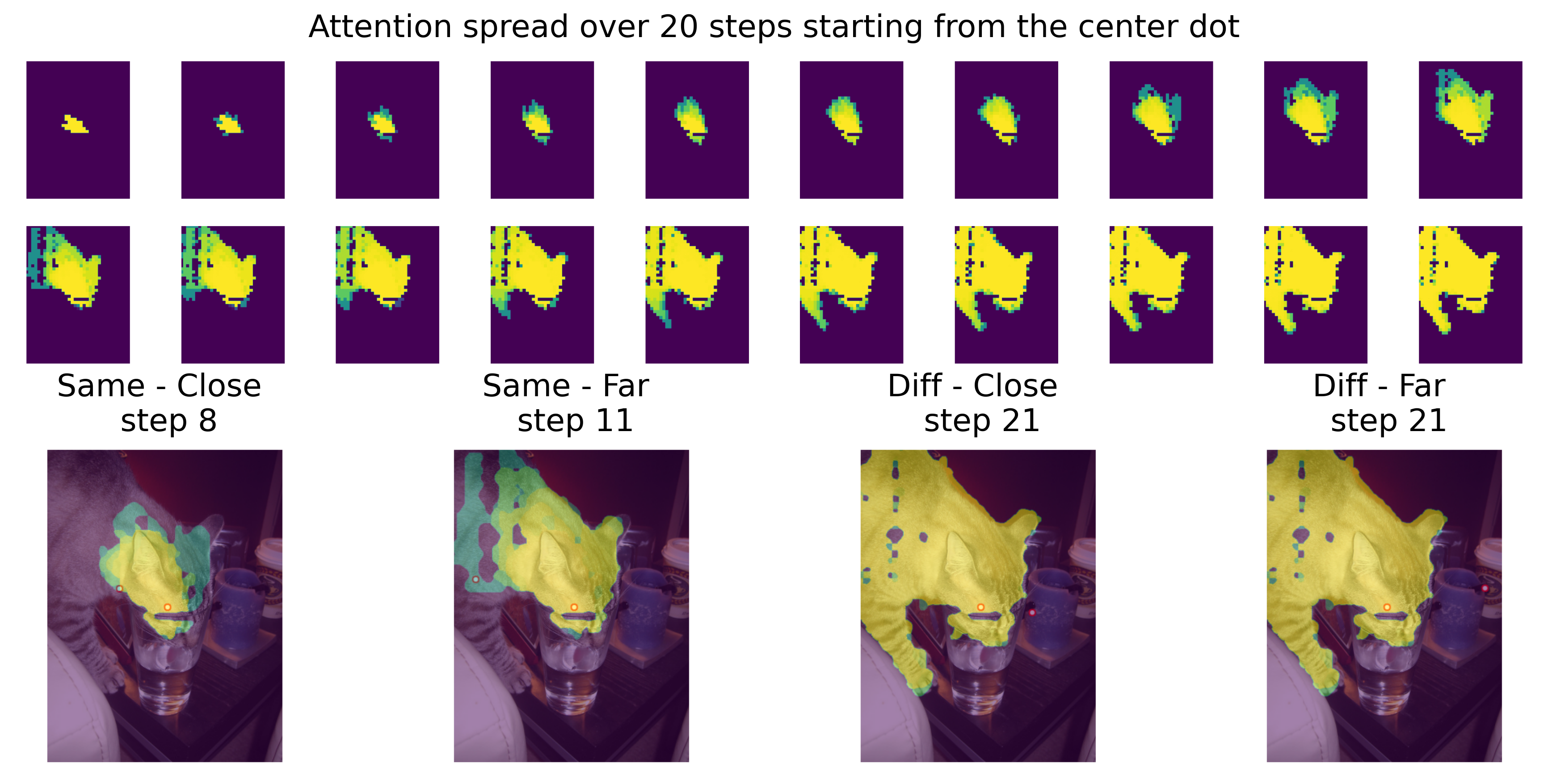}
\end{center}
\caption{20 steps of attention spreading from the center dot. Attention spread overlaid on the image for the steps that attention reached the peripheral dot in the same - close, same - far, different - close and different - far conditions shown at the bottom.}
\label{fig:spread_s5}
\end{figure*}

\begin{figure*}[h]
\begin{center}
\includegraphics[width=\linewidth]{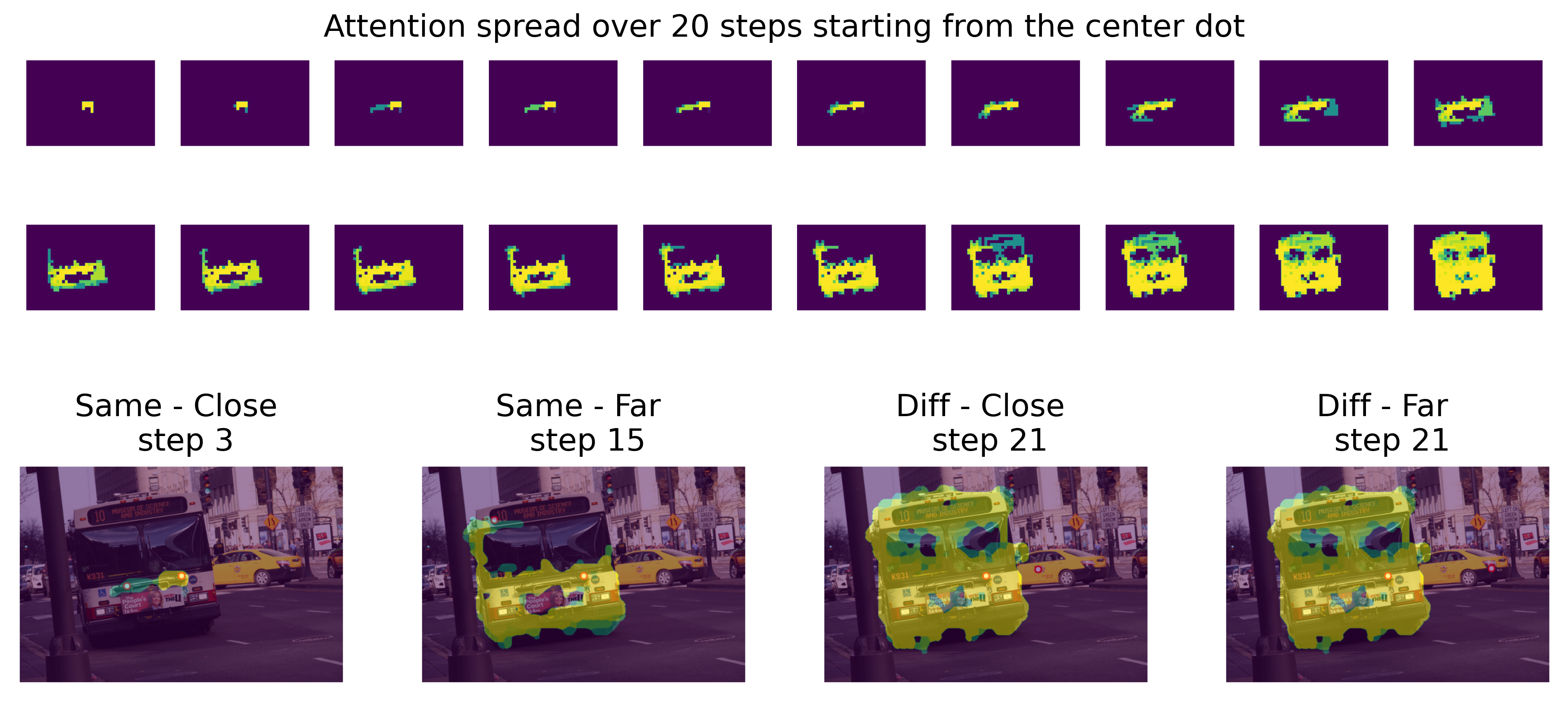}
\end{center}
\caption{20 steps of attention spreading from the center dot. Attention spread overlaid on the image for the steps that attention reached the peripheral dot in the same - close, same - far, different - close and different - far conditions shown at the bottom.}
\label{fig:spread_s6}
\end{figure*}

\begin{figure*}[h]
\begin{center}
\includegraphics[width=\linewidth]{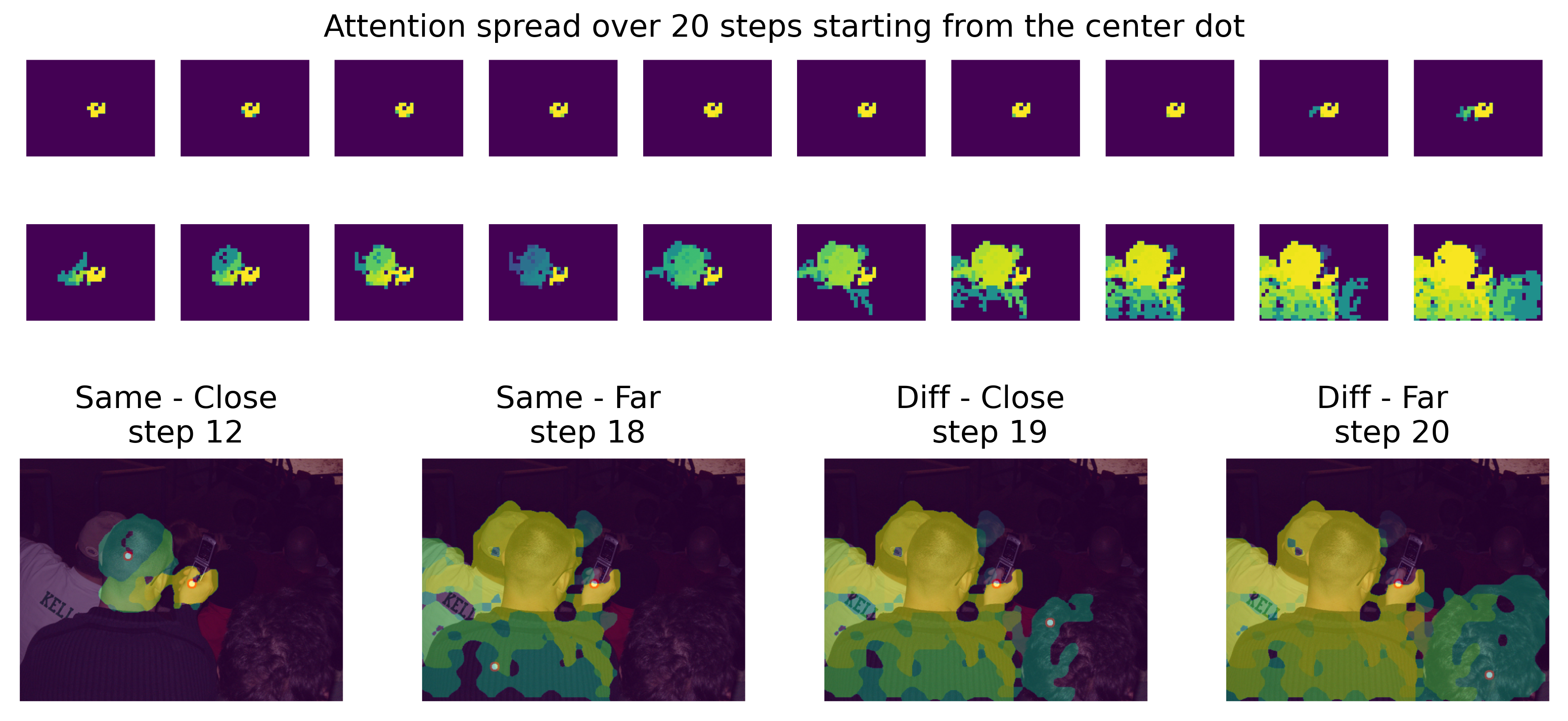}
\end{center}
\caption{20 steps of attention spreading from the center dot. Attention spread overlaid on the image for the steps that attention reached the peripheral dot in the same - close, same - far, different - close and different - far conditions shown at the bottom.}
\label{fig:spread_s7}
\end{figure*}

\begin{figure*}[h]
\begin{center}
\includegraphics[width=\linewidth]{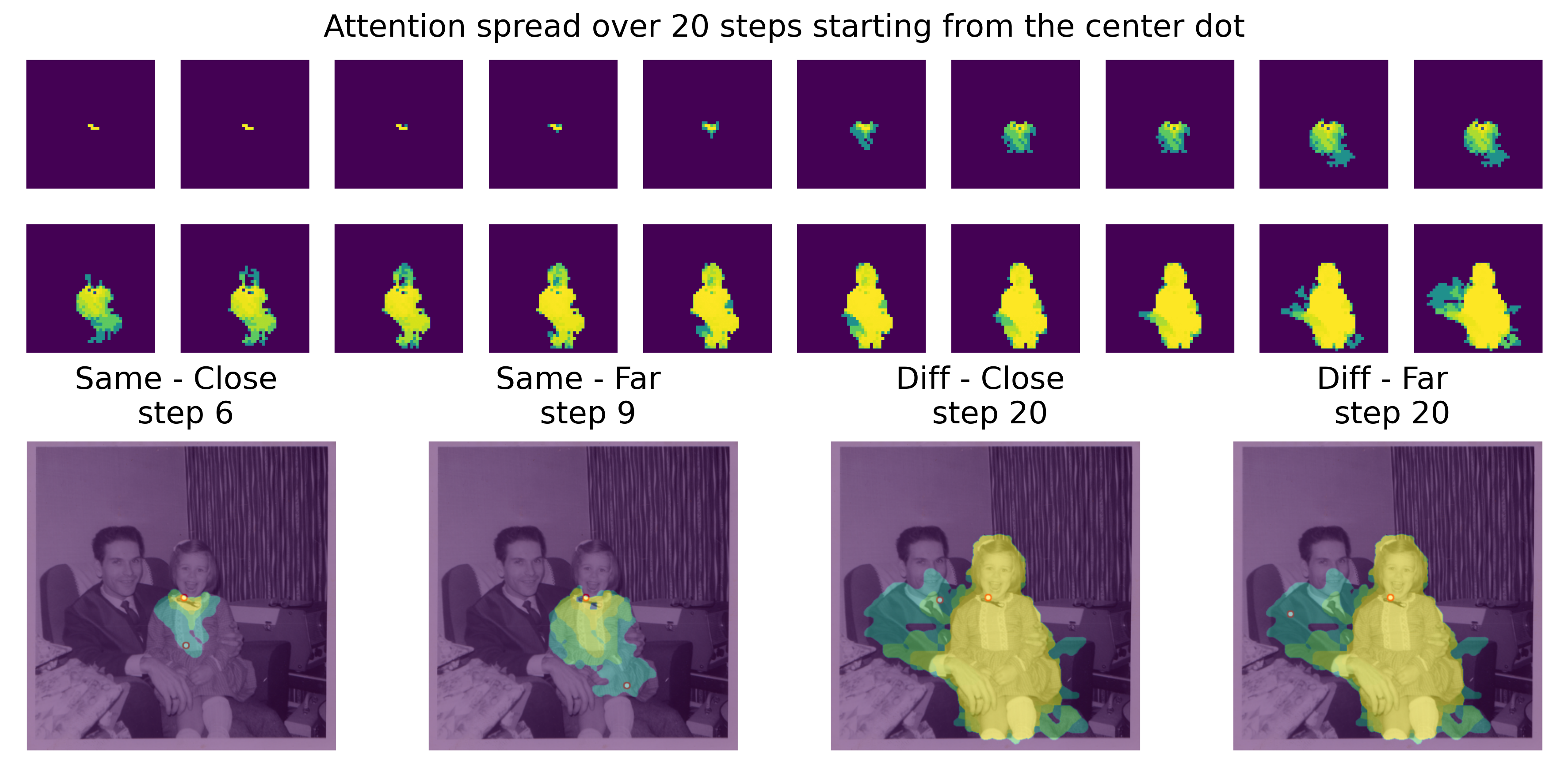}
\end{center}
\caption{20 steps of attention spreading from the center dot. Attention spread overlaid on the image for the steps that attention reached the peripheral dot in the same - close, same - far, different - close and different - far conditions shown at the bottom.}
\label{fig:spread_s8}
\end{figure*}

\begin{figure*}[h]
\begin{center}
\includegraphics[width=\linewidth]{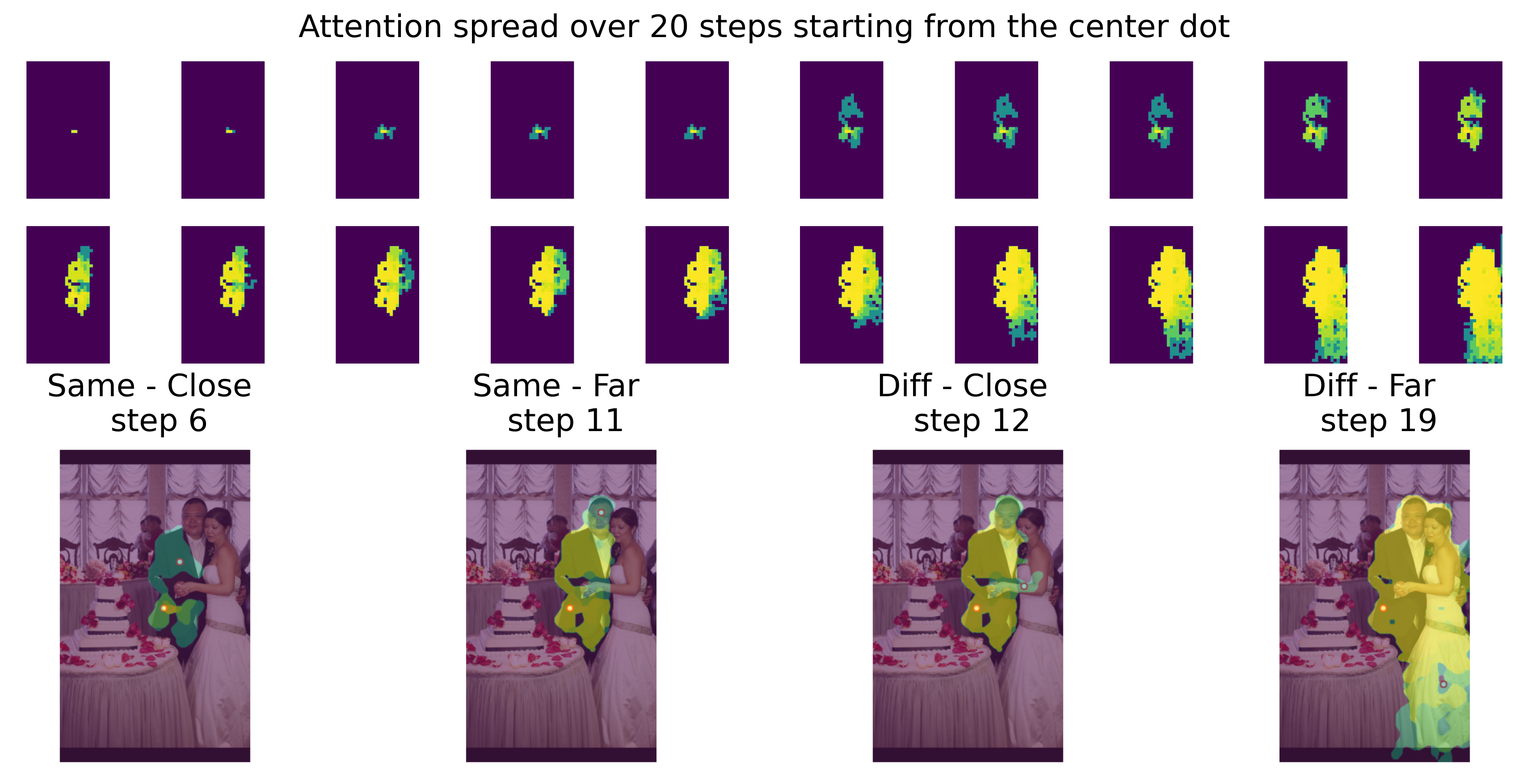}
\end{center}
\caption{20 steps of attention spreading from the center dot. Attention spread overlaid on the image for the steps that attention reached the peripheral dot in the same - close, same - far, different - close and different - far conditions shown at the bottom.}
\label{fig:spread_s9}
\end{figure*}

\begin{figure*}[h]
\begin{center}
\includegraphics[width=\linewidth]{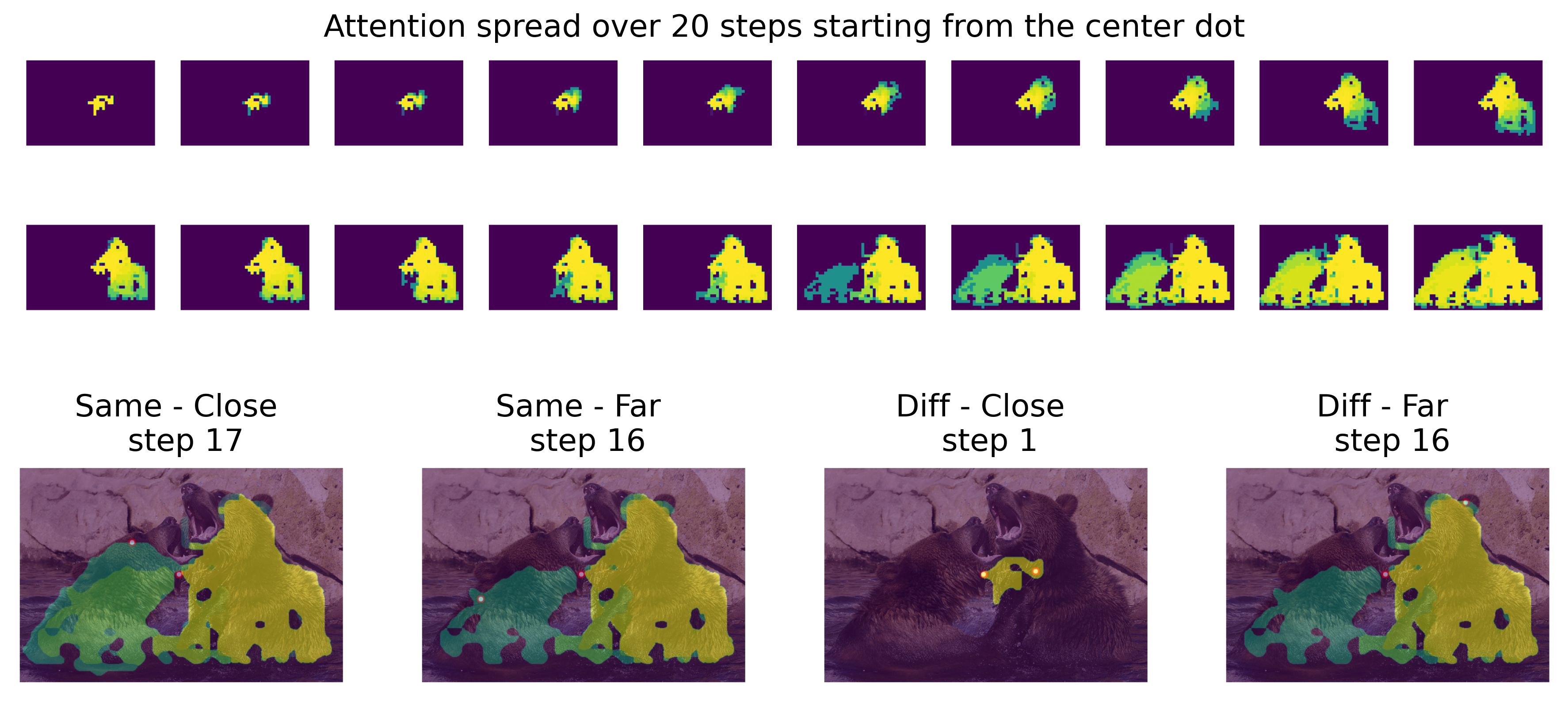}
\end{center}
\caption{20 steps of attention spreading from the center dot. Attention spread overlaid on the image for the steps that attention reached the peripheral dot in the same - close, same - far, different - close and different - far conditions shown at the bottom.}
\label{fig:spread_s10}
\end{figure*}

\begin{figure*}[h]
\begin{center}
\includegraphics[width=\linewidth]{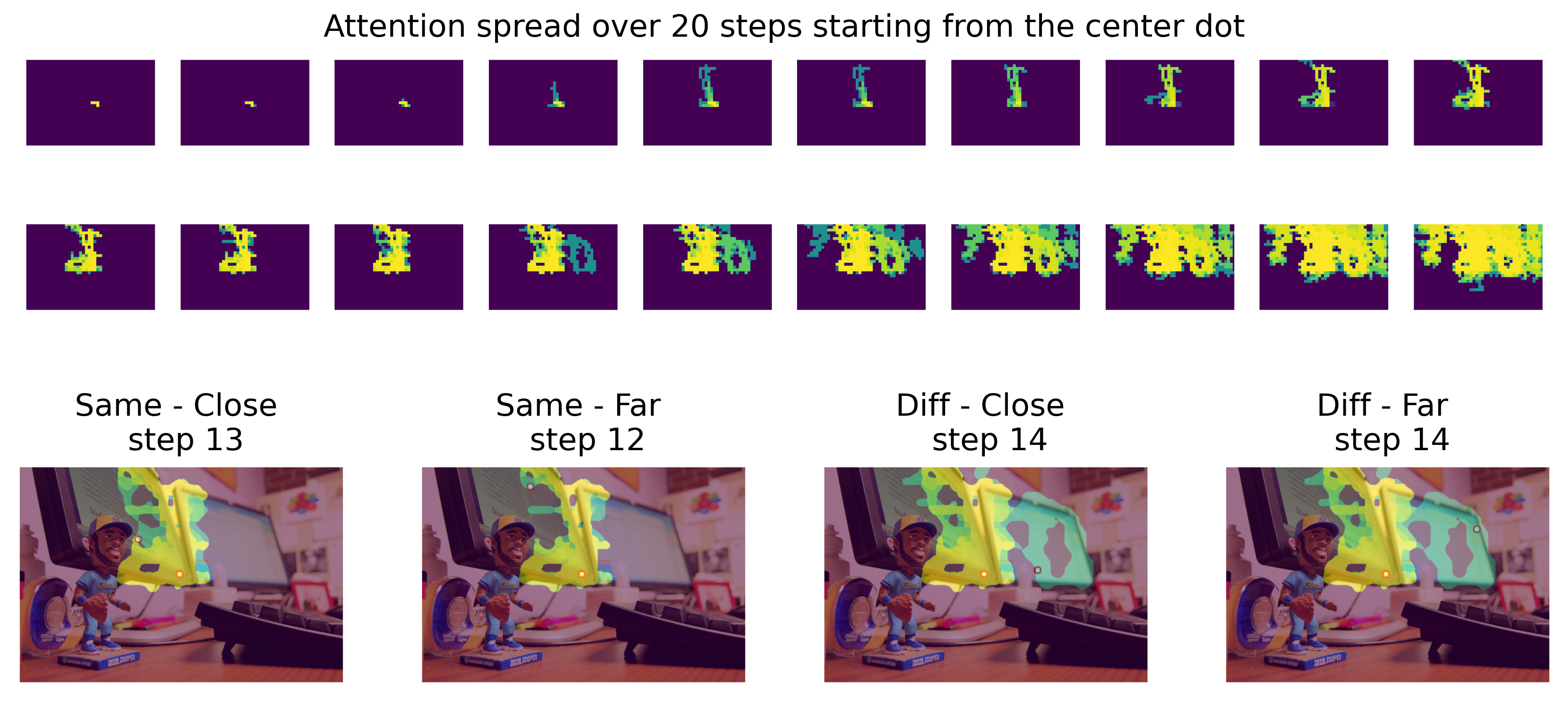}
\end{center}
\caption{20 steps of attention spreading from the center dot. Attention spread overlaid on the image for the steps that attention reached the peripheral dot in the same - close, same - far, different - close and different - far conditions shown at the bottom.}
\label{fig:spread_s11}
\end{figure*}

\begin{figure*}[h]
\begin{center}
\includegraphics[width=\linewidth]{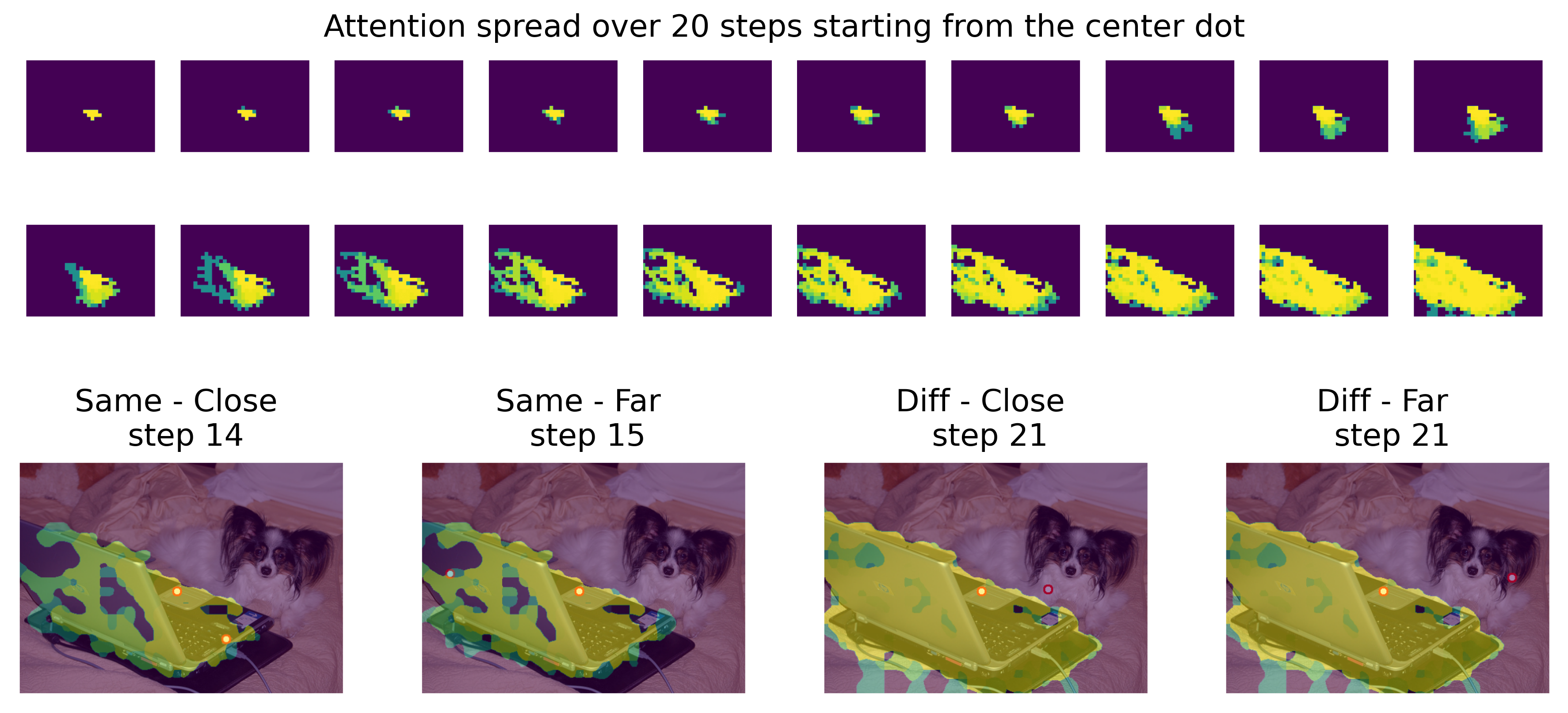}
\end{center}
\caption{20 steps of attention spreading from the center dot. Attention spread overlaid on the image for the steps that attention reached the peripheral dot in the same - close, same - far, different - close and different - far conditions shown at the bottom.}
\label{fig:spread_s12}
\end{figure*}

\begin{figure*}[h]
\begin{center}
\includegraphics[width=\linewidth]{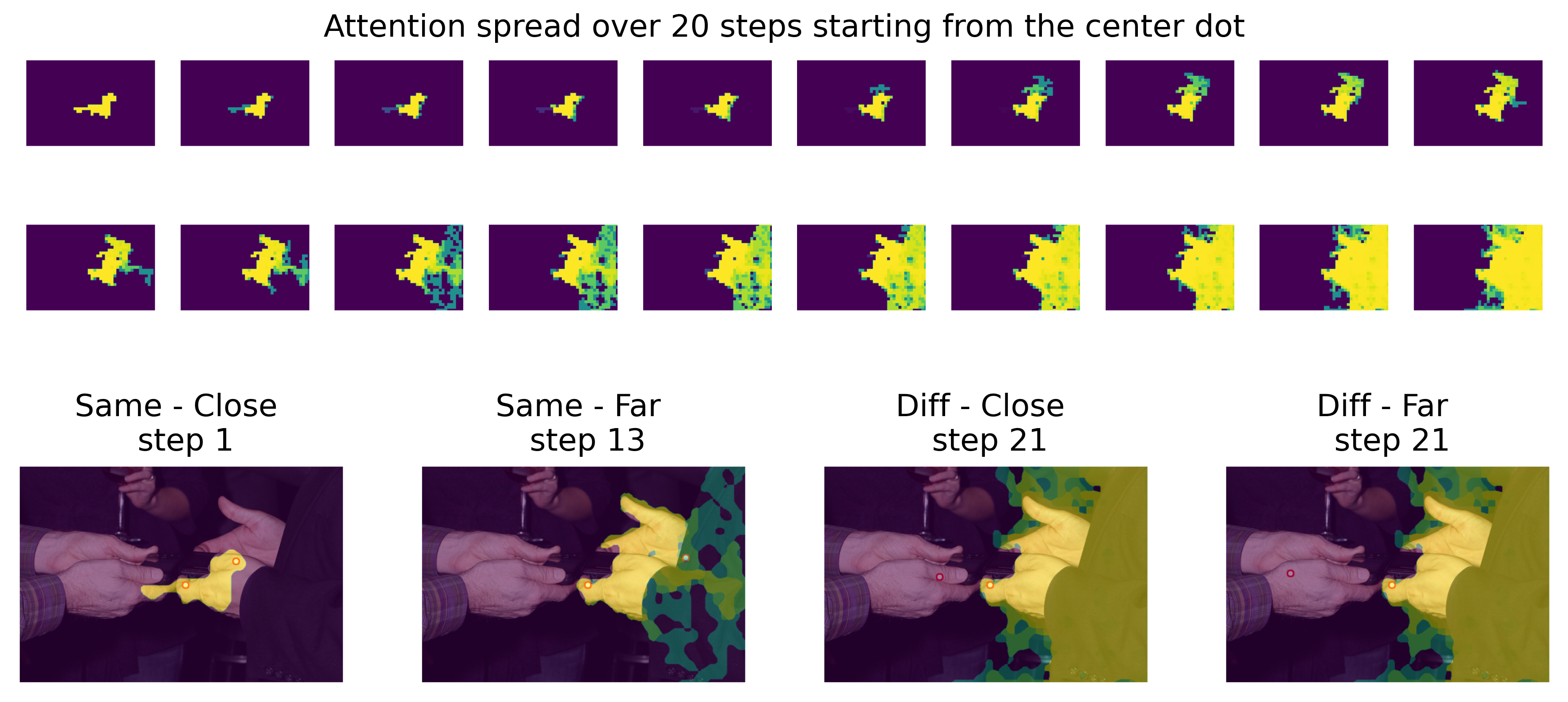}
\end{center}
\caption{20 steps of attention spreading from the center dot. Attention spread overlaid on the image for the steps that attention reached the peripheral dot in the same - close, same - far, different - close and different - far conditions shown at the bottom.}
\label{fig:spread_s13}
\end{figure*}

\begin{figure*}[h]
\begin{center}
\includegraphics[width=\linewidth]{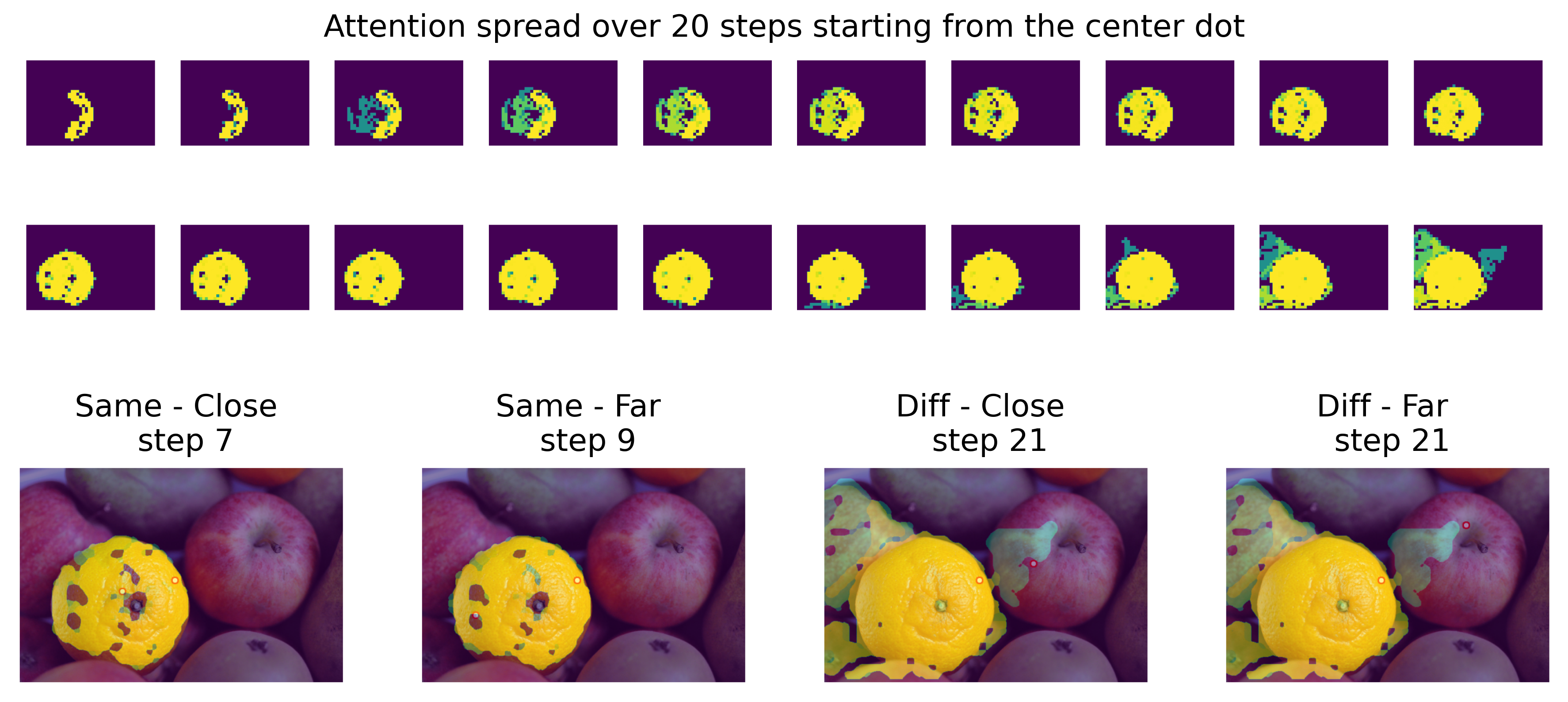}
\end{center}
\caption{20 steps of attention spreading from the center dot. Attention spread overlaid on the image for the steps that attention reached the peripheral dot in the same - close, same - far, different - close and different - far conditions shown at the bottom.}
\label{fig:spread_s14}
\end{figure*}

\begin{figure*}[h]
\begin{center}
\includegraphics[width=\linewidth]{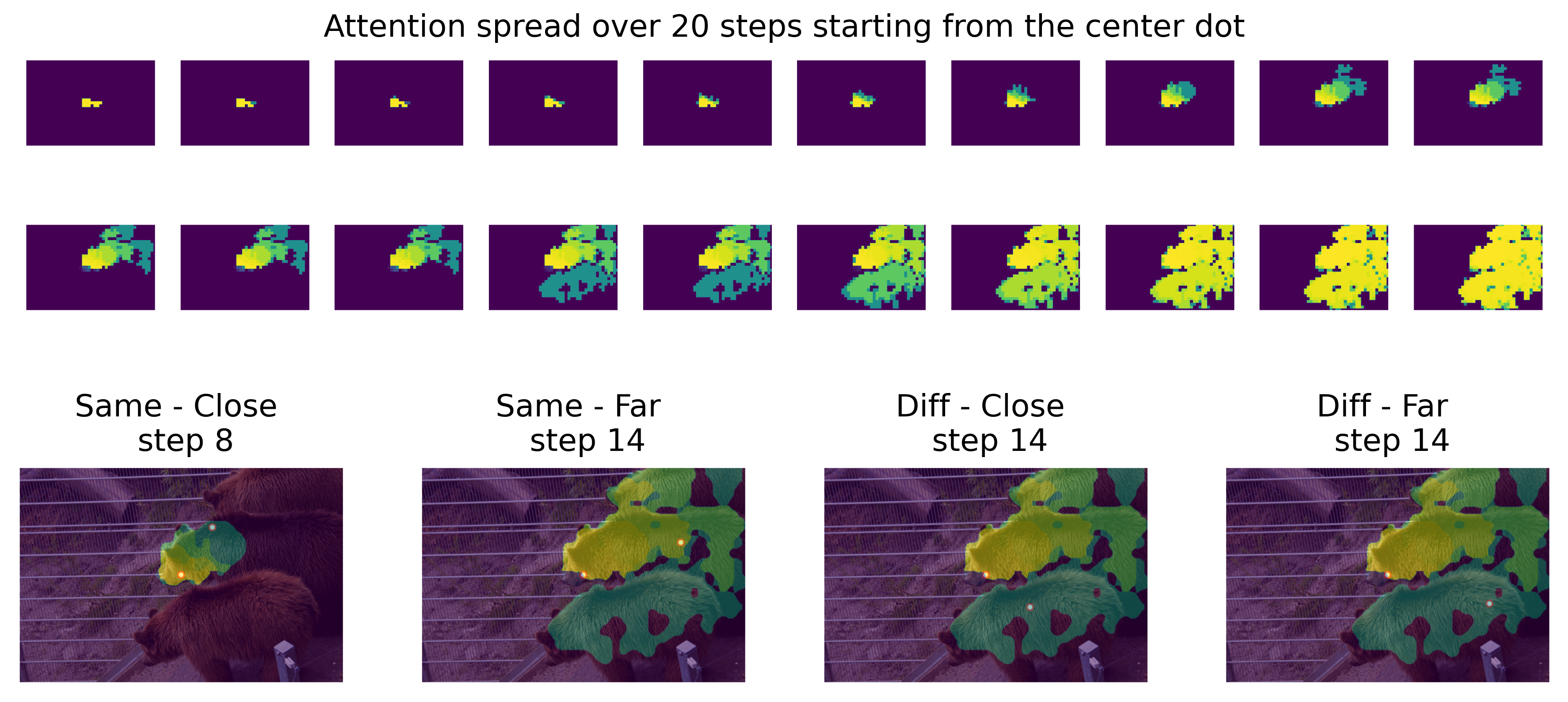}
\end{center}
\caption{20 steps of attention spreading from the center dot. Attention spread overlaid on the image for the steps that attention reached the peripheral dot in the same - close, same - far, different - close and different - far conditions shown at the bottom.}
\label{fig:spread_s15}
\end{figure*}

\begin{figure*}[h]
\begin{center}
\includegraphics[width=\linewidth]{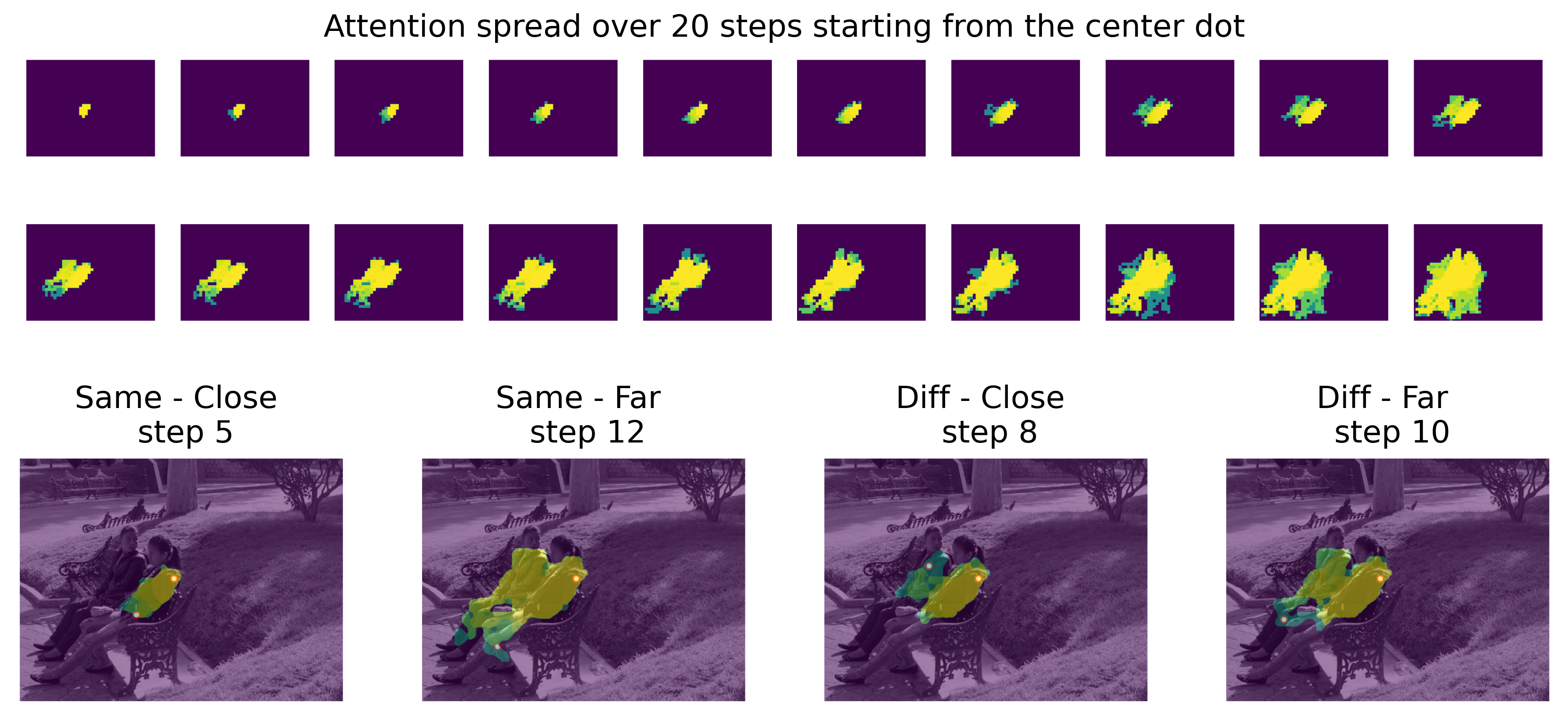}
\end{center}
\caption{20 steps of attention spreading from the center dot. Attention spread overlaid on the image for the steps that attention reached the peripheral dot in the same - close, same - far, different - close and different - far conditions shown at the bottom.}
\label{fig:spread_s16}
\end{figure*}

\begin{figure*}[h]
\begin{center}
\includegraphics[width=\linewidth]{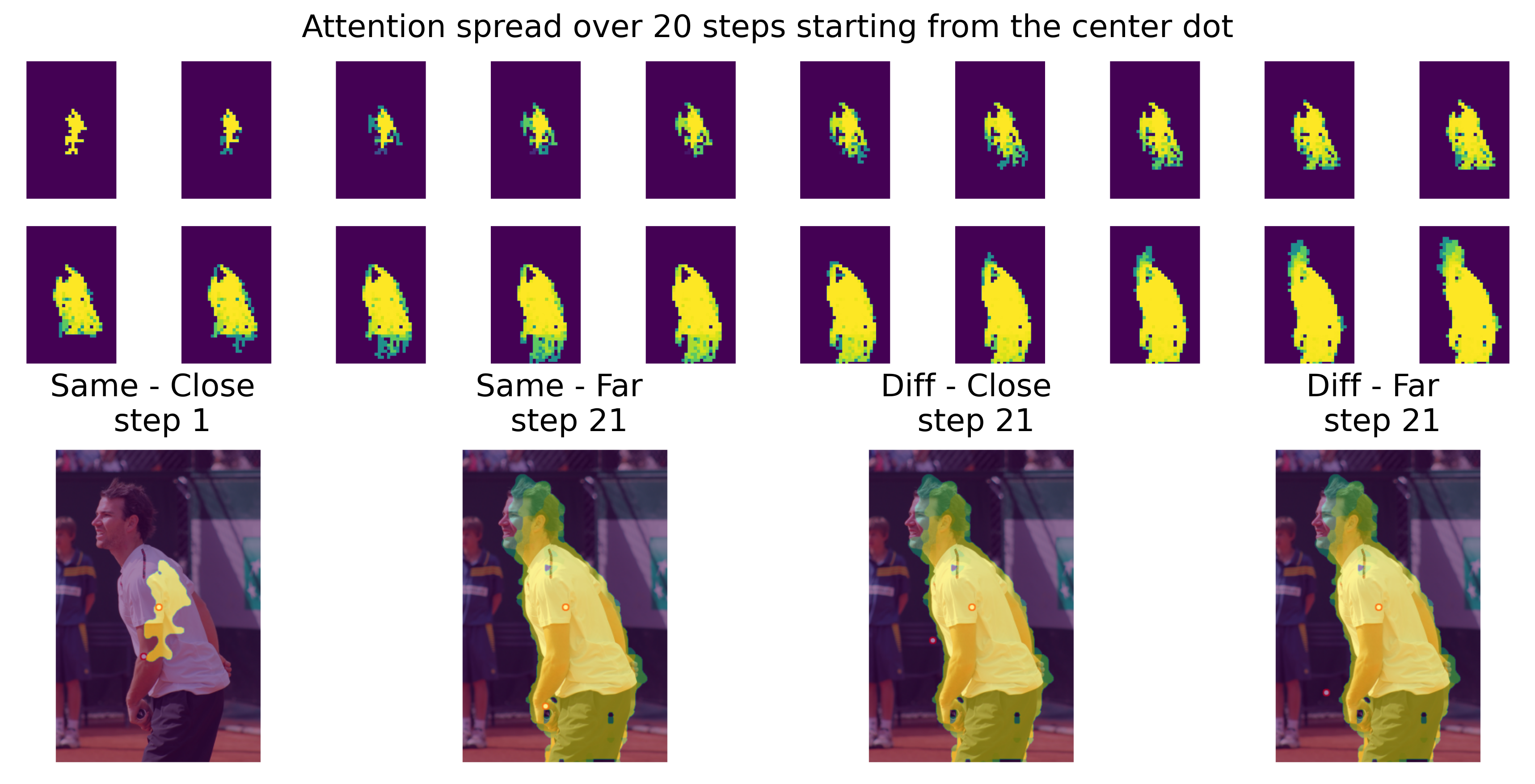}
\end{center}
\caption{20 steps of attention spreading from the center dot. Attention spread overlaid on the image for the steps that attention reached the peripheral dot in the same - close, same - far, different - close and different - far conditions shown at the bottom.}
\label{fig:spread_s17}
\end{figure*}

\begin{figure*}[h]
\begin{center}
\includegraphics[width=\linewidth]{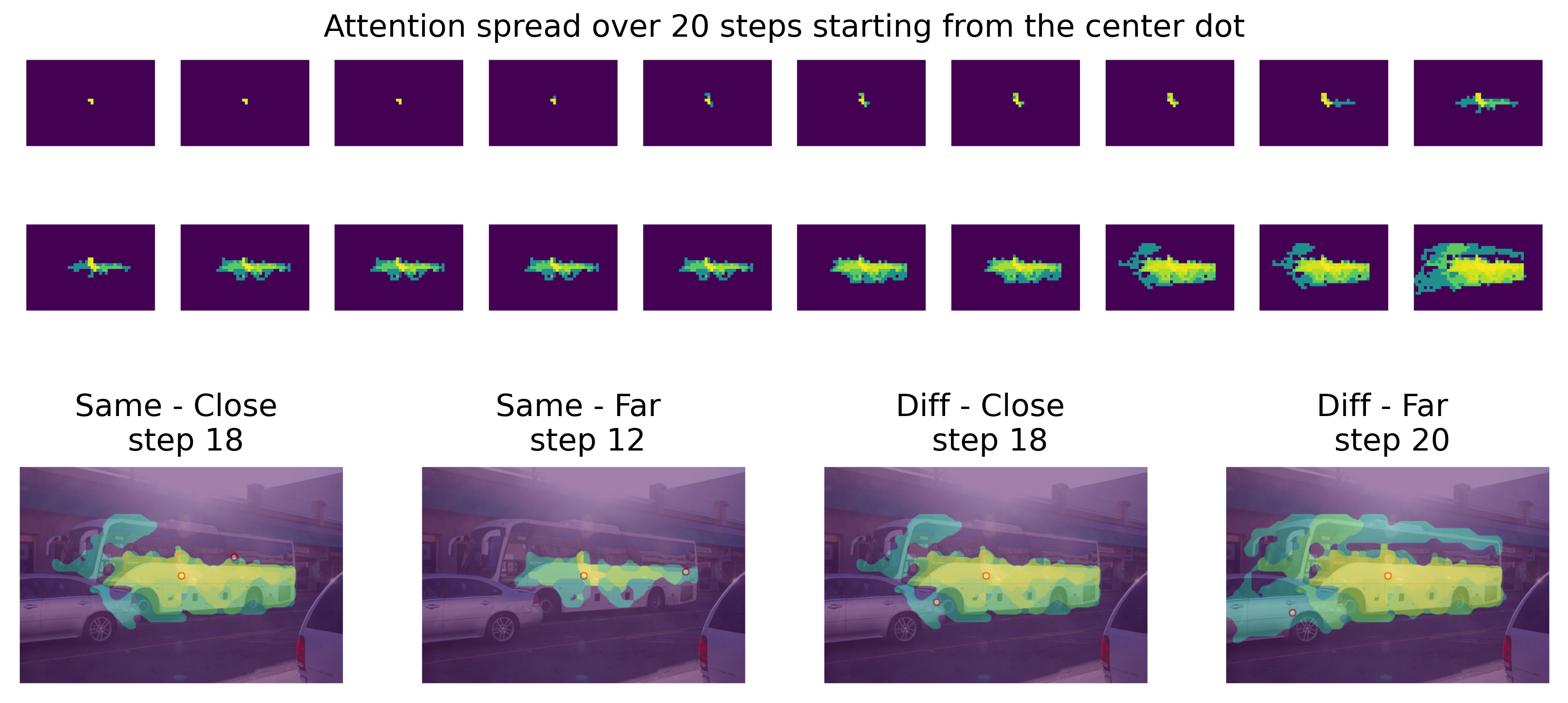}
\end{center}
\caption{20 steps of attention spreading from the center dot. Attention spread overlaid on the image for the steps that attention reached the peripheral dot in the same - close, same - far, different - close and different - far conditions shown at the bottom.}
\label{fig:spread_s18}
\end{figure*}

\begin{figure*}[h]
\begin{center}
\includegraphics[width=\linewidth]{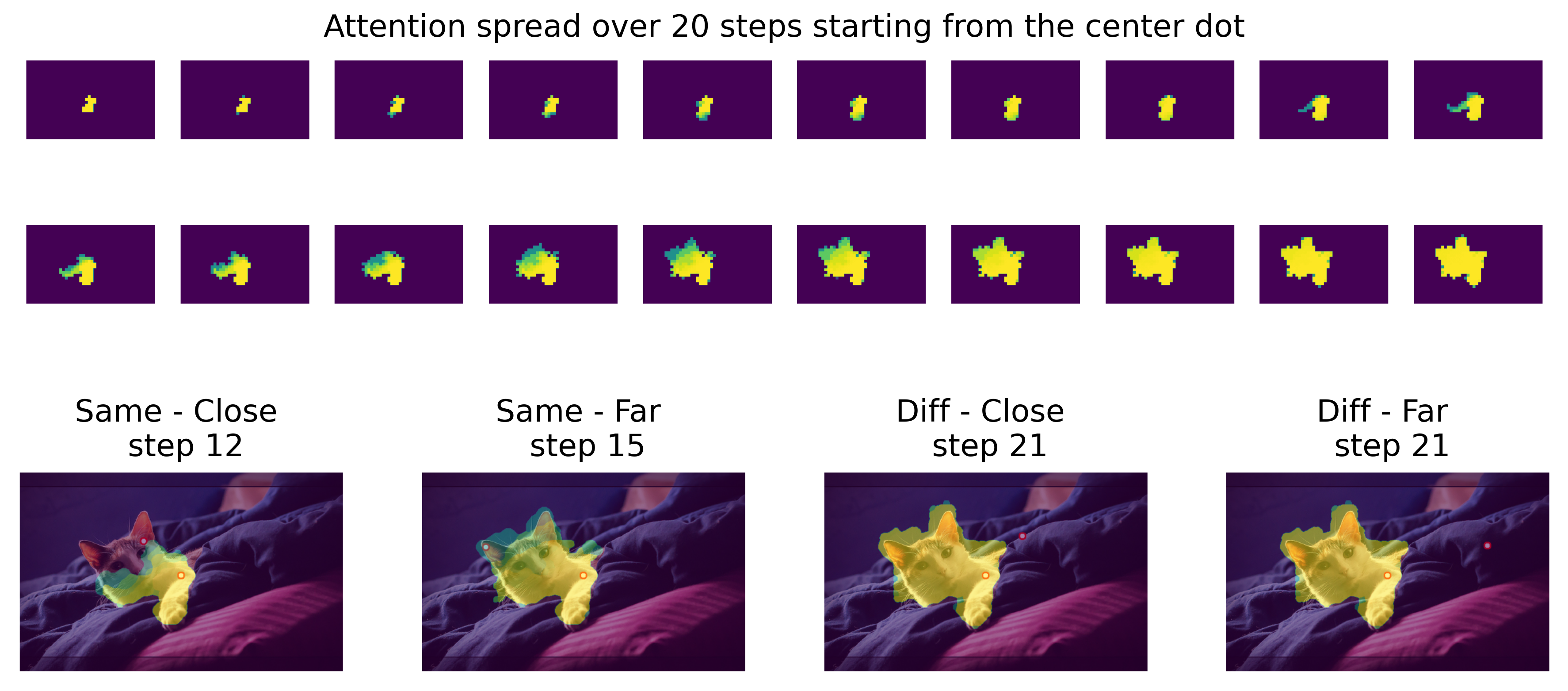}
\end{center}
\caption{20 steps of attention spreading from the center dot. Attention spread overlaid on the image for the steps that attention reached the peripheral dot in the same - close, same - far, different - close and different - far conditions shown at the bottom.}
\label{fig:spread_s19}
\end{figure*}

\begin{figure*}[h]
\begin{center}
\includegraphics[width=\linewidth]{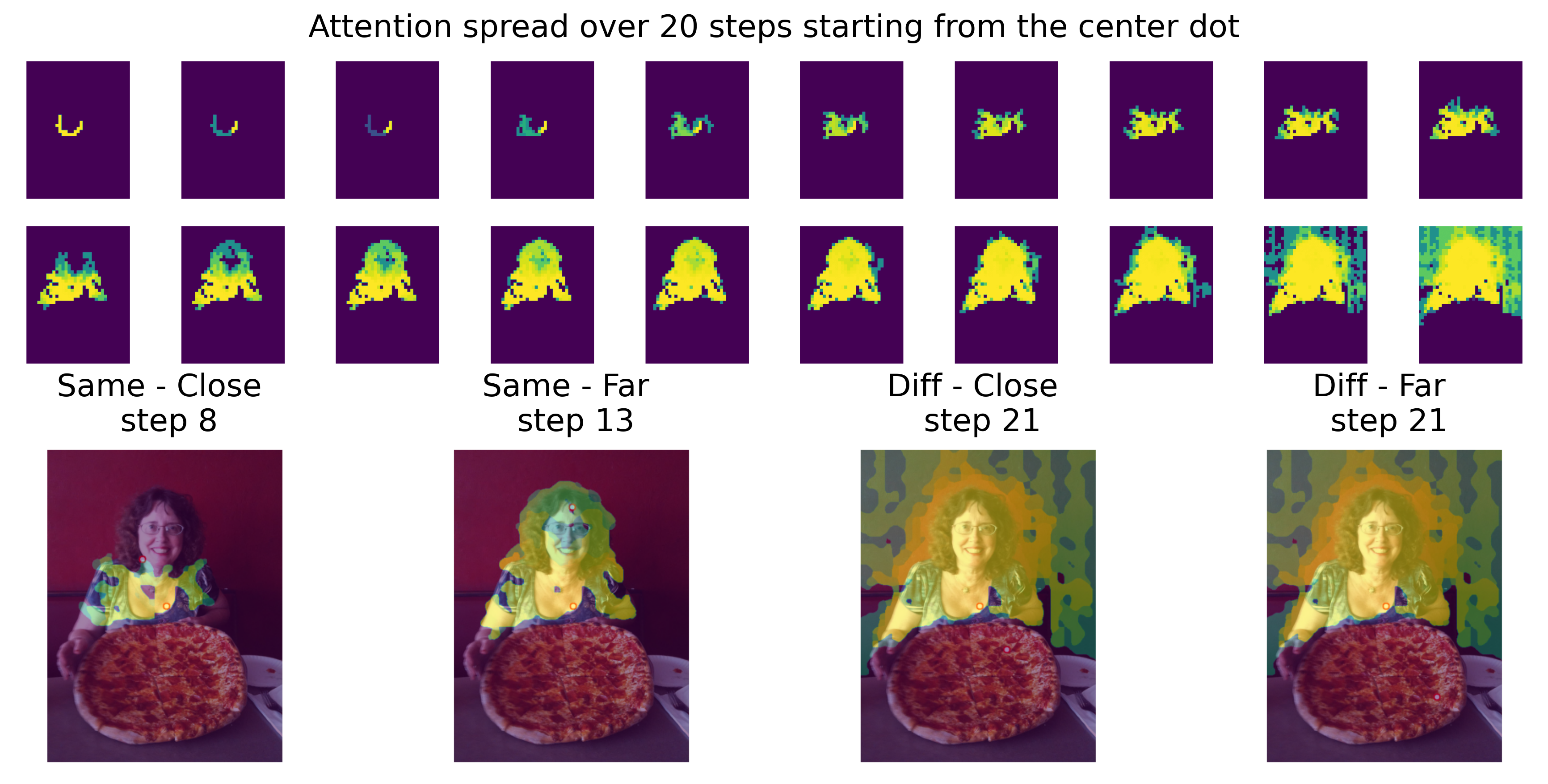}
\end{center}
\caption{20 steps of attention spreading from the center dot. Attention spread overlaid on the image for the steps that attention reached the peripheral dot in the same - close, same - far, different - close and different - far conditions shown at the bottom.}
\label{fig:spread_slast}
\end{figure*}

Fig.~\ref{fig:spread_s1} through Fig.~\ref{fig:spread_slast} show more example model outputs. 

\begin{table}[h]
\caption{AUCs for all model runs}
\label{tab:all_runs_roc}
    \centering
\def\arraystretch{1.9}
\begin{tabular}{|p{3cm}||p{1.4cm}|p{1.7cm}|p{1.cm}|p{1cm}|p{1cm}|p{1cm}| }
 \hline
  run name & training objective  & architecture & model size & patch size & feature type & AUC \\
 \hline
 DINOv2\_ViTb14 & DINO v2& ViT & base & 14 & key & 0.870\\
  & & &  &  & query & 0.877\\
  & & &  &  & value & 0.770\\
 \hline
 DINOv2\_ViTl14 & DINO v2& ViT & large & 14 & key & 0.856\\
  & & &  &  & query & 0.854\\
  & & &  &  & value &0.785\\
 \hline
 DINOv2\_ViTg14 & DINO v2& ViT & giant & 14 & key & 0.810\\
  & & &  &  & query & 0.840\\
  & & &  &  & value & 0.770\\
 \hline
 DINO\_ViTb16 & DINO & ViT & base & 16 & key & 0.800\\
  & & &  &  & query & 0.800\\
  & & &  &  & value & 0.786\\
 \hline
 DINO\_ViTb8 & DINO & ViT & base & 8 & key & 0.768\\
  & & &  &  & query & 0.785\\
  & & &  &  & value & 0.791\\
 \hline
 MAE\_ViTb16 & MAE & ViT & base & 16 & key & 0.840\\
  & & &  &  & query & 0.850\\
  & & &  &  & value &0.820\\
 \hline
 DINO\_ResNet50 & DINO & ResNet50 & & & conv & 0.618\\
 \hline
 ImageNet\_ResNet50 & ImageNet & ResNet50 &  &  & conv & 0.611\\
 \hline
\end{tabular}
    \centering
\end{table}
\vspace*{2mm}

\begin{table}[h]
\caption{Human-model correlation for all model runs}
\label{tab:all_runs_corr}
    \centering
\def\arraystretch{2}
\begin{tabular}{|p{3cm}||p{1.4cm}|p{1.7cm}|p{1.cm}|p{1cm}|p{1cm}|p{1.5cm}| }
 \hline
  run name & training objective  & architecture & model size & patch size & feature type & Spearman rank    correlation \\
 \hline
 DINOv2\_ViTb14 & DINO v2& ViT & base & 14 & key & 0.238\\
  & & &  &  & query & 0.252\\
  & & &  &  & value & 0.213\\
 \hline
 DINOv2\_ViTl14 & DINO v2& ViT & large & 14 & key & 0.241\\
  & & &  &  & query & 0.231\\
  & & &  &  & value &0.219\\
 \hline
 DINOv2\_ViTg14 & DINO v2& ViT & giant & 14 & key & 0.195\\
  & & &  &  & query & 0.206\\
  & & &  &  & value & 0.189\\
 \hline
 DINO\_ViTb16 & DINO & ViT & base & 16 & key & 0.201\\
  & & &  &  & query & 0.215\\
  & & &  &  & value & 0.212\\
 \hline
 DINO\_ViTb8 & DINO & ViT & base & 8 & key & 0.187\\
  & & &  &  & query & 0.229\\
  & & &  &  & value & 0.227\\
 \hline
 MAE\_ViTb16 & MAE & ViT & base & 16 & key & 0.218\\
  & & &  &  & query & 0.234\\
  & & &  &  & value &0.217\\
 \hline
 DINO\_ResNet50 & DINO & ResNet50 & & & conv & 0.116\\
 \hline
 ImageNet\_ResNet50 & ImageNet & ResNet50 &  &  & conv & 0.103\\
 \hline
\end{tabular}
    \centering
\end{table}
\vspace*{2mm}

\end{document}